\documentclass{article}

\usepackage{microtype}
\usepackage{graphicx}
\usepackage{subcaption}
\usepackage{booktabs} 
\usepackage{placeins}
\usepackage{hyperref}

\usepackage[accepted]{icml2026}

\usepackage{amsmath}
\usepackage{amssymb}
\usepackage{mathtools}
\usepackage{amsthm}

\usepackage{booktabs}
\usepackage{tabularx}
\usepackage{array}
\usepackage{multirow}
\usepackage{hyperref}

\newcolumntype{Y}{>{\raggedright\arraybackslash}X}

\usepackage[capitalize,noabbrev]{cleveref}
\usepackage{hyperref}
\usepackage{url}
\usepackage{graphicx}
\usepackage{booktabs}  
\usepackage{pifont}    
\usepackage{amssymb}   
\usepackage[dvipsnames]{xcolor}
\usepackage{float}
\theoremstyle{plain}

\theoremstyle{definition}

\theoremstyle{remark}

\usepackage[textsize=tiny]{todonotes}

\icmltitlerunning{PyHealth 2.0: A Comprehensive Clinical Deep Learning Toolkit}

\begin{document}

\twocolumn[
  \icmltitle{PyHealth 2.0: A Comprehensive Open-Source Toolkit for Accessible and Reproducible Clinical Deep Learning}



  \icmlsetsymbol{equal}{*}

\begin{icmlauthorlist}
    \icmlauthor{John Wu}{uiuc,pyhealth,keiji}
    \icmlauthor{Yongda Fan}{uiuc,pyhealth}
    \icmlauthor{Zhenbang Wu}{uiuc}
    \icmlauthor{Paul Landes}{uiuc-med}
    \icmlauthor{Eric Schrock}{uiuc,pyhealth}
    \icmlauthor{Sayeed Sajjad Razin}{pyhealth,buet}
    \icmlauthor{Arjun Chatterjee}{uiuc,pyhealth}
    \icmlauthor{Naveen Baskaran}{uiuc,pyhealth}
    \icmlauthor{Joshua Steier}{pyhealth}
    \icmlauthor{Andrea Fitzpatrick}{uiuc,pyhealth}
    \icmlauthor{Bilal Arif}{uiuc,pyhealth}
    \icmlauthor{Rian Atri}{pyhealth}
    \icmlauthor{Jathurshan Pradeepkumar}{uiuc,pyhealth}
    \icmlauthor{Siddhartha Laghuvarapu}{uiuc,pyhealth}
    \icmlauthor{Junyi Gao}{edinburgh,hdruk}
    \icmlauthor{Adam R. Cross}{uiuc-med}
    \icmlauthor{Jimeng Sun}{uiuc,keiji}
\end{icmlauthorlist}

\icmlaffiliation{uiuc}{University of Illinois Urbana-Champaign, Urbana, IL, USA}
\icmlaffiliation{pyhealth}{PyHealth Research Initiative}
\icmlaffiliation{keiji}{Keiji AI}
\icmlaffiliation{uiuc-med}{University of Illinois College of Medicine, Chicago, IL, USA}
\icmlaffiliation{edinburgh}{The University of Edinburgh, Edinburgh, UK}
\icmlaffiliation{hdruk}{Health Data Research UK, London, UK}
\icmlaffiliation{buet}{Department of Biomedical Engineering, Bangladesh University of Engineering and Technology, Dhaka 1000, Bangladesh}

\icmlcorrespondingauthor{John Wu}{johnwu3@illinois.edu}

  \icmlkeywords{Machine Learning, ICML}

  \vskip 0.3in
]



\printAffiliationsAndNotice{}  

\begin{abstract}
Difficulty replicating baselines, high computational costs, and required domain expertise create persistent barriers to clinical AI research. To address these challenges, we introduce PyHealth 2.0, an enhanced clinical deep learning toolkit that enables predictive modeling in as few as 7 lines of code. PyHealth 2.0 offers three key contributions: (1) a comprehensive toolkit addressing reproducibility and compatibility challenges by unifying 15+ datasets, 20+ clinical tasks, 25+ models, 5+ interpretability methods, and 5+ uncertainty quantification methods within a single framework that supports diverse clinical data modalities—signals, text, imaging, and electronic health records—with translation of 5+ medical coding standards; (2) accessibility-focused design accommodating multimodal data and diverse computational resources with up to 39× faster processing and 20× lower memory usage, enabling work from 16GB laptops to production systems; and (3) an active open-source community of 400+ members lowering domain expertise barriers through extensive documentation, reproducible research contributions, and collaborations with academic health systems and industry partners, including multi-language support via RHealth. PyHealth 2.0 establishes an open-source foundation and community advancing accessible, reproducible healthcare AI. Project details at \url{https://pyhealth.dev/}.

\end{abstract}


\section{Introduction} \label{sec:intro}
Regardless of the task, clinical deep learning models functionally follow a five-step pipeline: data processing, machine learning task definition, model initialization, training, and evaluation \citep{janiesch2021machine,shinde2018review_dl_apps,wang2024recent_survey_clinical_predictive_modeling}. However, despite this standardized modeling process, reproducibility remains an increasingly difficult challenge due to both the rapid progress in AI developments and the lack of available, operable code \citep{mcdermott2019reproducibilitymachinelearninghealth,mcdermott2021reproducibility,beam2020challenges_reproducibility}. If any step is missing from this pipeline, reported results from clinical predictive models become practically irreproducible, making further development difficult. As such, it is crucial that all pipeline steps are available and reproducible.

\textbf{The Reproducibility Crisis.} Many clinical predictive models use similar data modalities—structured diagnosis and procedure codes, lab events—meaning their data processing often contains identical steps regardless of the task \citep{wang2024recent_survey_clinical_predictive_modeling}. Nonetheless, researchers consistently implement their own processing approaches with variations in implementation and random seeds, making reproducibility challenging \citep{vandewater2024icubenchmarkflexiblemulticenter_YAIB, pmlr-v68-johnson17a_reprod_mortality_pred}. Furthermore, since each set of reported results is often associated with its own repository, directly auditing claims becomes tedious and unrewarding. When investigated, the majority of cohorts used in ML pipelines were found irreproducible \citep{pmlr-v68-johnson17a_reprod_mortality_pred}. Standardizing and centralizing these repeatable steps is key to improving pipeline reproducibility and transparency.

\textbf{Dependency and Compatibility Challenges.} Repositories often rely on dependencies that are defunct or incompatible with existing work environments \citep{hassan2024reproducibility_dependency, semmelrock2023reproducibilitymachinelearningdrivenresearch}. While Docker containers provide explicit research replication \citep{boettiger2015introduction_docker_reproducibility}, they can be cumbersome, requiring additional dependencies and engineering practices beyond simple Python coding. Moreover, the goal of reproducibility extends beyond replication to exploring methods for developing better clinical AI models. A more reproducible future requires a unified framework of tested software compatible with standard working environments.

\textbf{Multimodal Complexity and Computational Barriers.} In recent years, the number of modalities considered within clinical AI systems has drastically increased, as electronic health records (EHR) are intrinsically multimodal \citep{acosta2022multimodal_biomed}. From signals and images to lab events and codes, a unified framework must adapt to numerous differing modalities, as patient profiles are fundamentally multimodal. Large EHR datasets such as MIMIC-IV contain over 100 million lab events \citep{johnson2023mimic4}, resulting in memory usage beyond conventional workstations. Our findings show memory requirements can balloon to several hundred gigabytes of RAM, making work on large clinical datasets highly infeasible on typical machines. Making healthcare AI more accessible requires reducing memory requirements for model training to fit within consumer-grade hardware.

\textbf{The Domain Knowledge Gap.} Finally, prerequisite domain knowledge creates barriers to auditing and understanding whether approaches are truly clinically relevant. As existing surveys evidence, a gap persists between the traditional machine learning and clinical communities \citep{nissar2023bridging_gap_between_tech_and_medicine}. The machine learning tasks developed should be crucial toward improving performance on problems highly relevant to real-world clinical needs. Achieving this requires not only technical expertise but also specific clinical experience to define available features and validate model performance. Bridging this cultural and technical gap between clinical experts and experienced AI researchers is crucial to ensure valuable human and computational resources are not wasted on the wrong problems.

\begin{figure*}[h]
    \centering
    \includegraphics[width=1.0\textwidth]{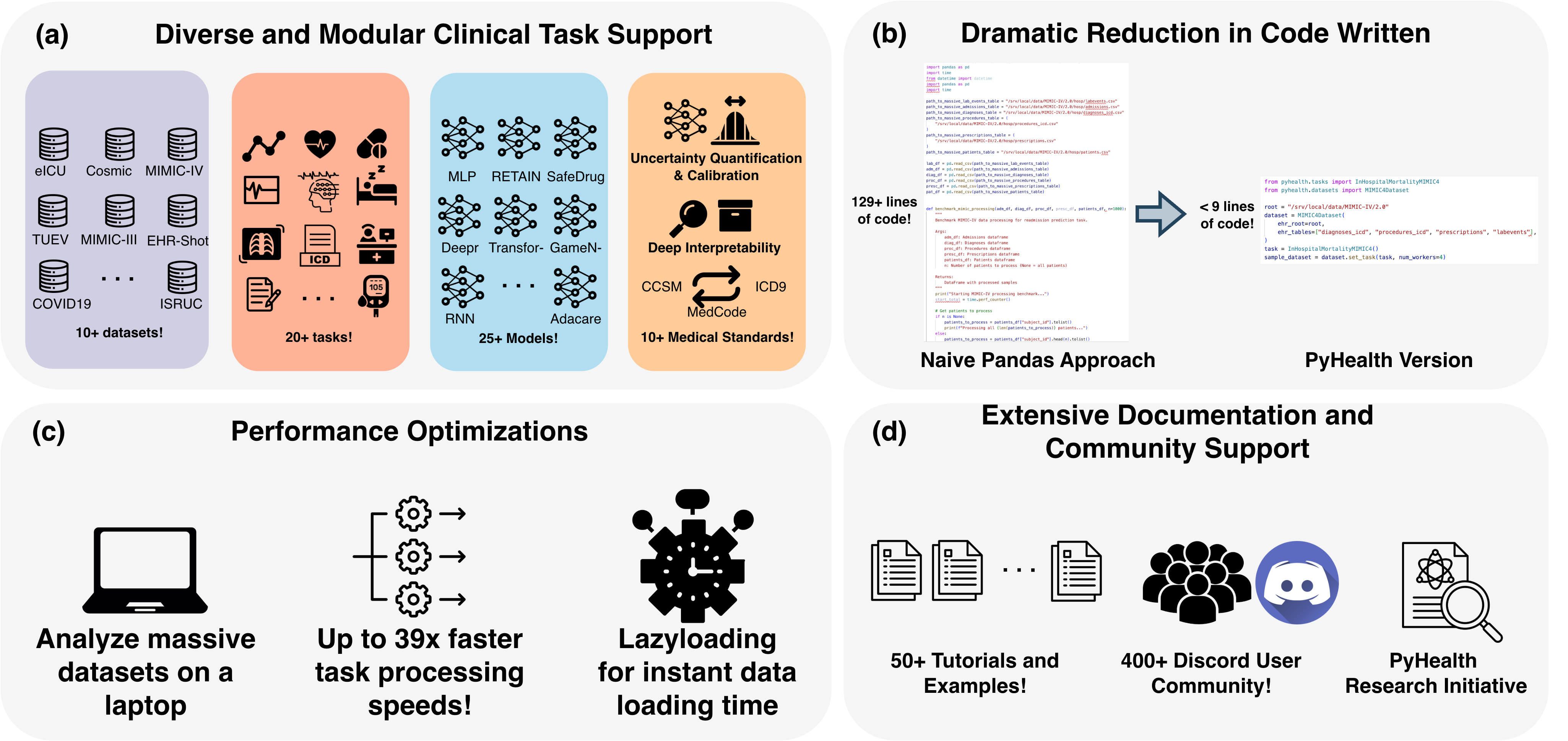}
    \caption{\textbf{PyHealth as a comprehensive healthcare AI development toolkit.} To tackle the reproducibility crisis in healthcare AI, (a) PyHealth has re-implemented over 10 different datasets for data loading, several dozen tasks, and over 25 different models with a variety of post-hoc model deployment features for better evaluating model performance. By re-implementing many of these components with standardized interfaces, (b) the amount of code required to generate samples ready for training on a specific ML task has been reduced from over 129 lines to fewer than 9. To enable users with limited computational resources, (c) PyHealth contains a variety of backend optimizations that enable data processing on modern laptops and support up to 39× faster processing speeds than alternatives (Figure \ref{fig:pyhealth_performance}). Finally, as a growing open-source community, (d) PyHealth 2.0 further embodies reproducible research principles through extensive examples and collaborative development.}
    \label{fig:pyhealth_benefits}
\end{figure*}

\textbf{Our Contributions (Figure \ref{fig:pyhealth_benefits}).} We introduce PyHealth 2.0, a deep learning toolkit which offers direct solutions to each of these problems through three key contributions: (1) \textbf{a comprehensive toolkit} spanning datasets, models, tasks, and evaluation methods that bridges the gap between technical and clinical domains through standardized implementations addressing reproducibility challenges; (2) \textbf{accessibility-focused design} that accommodates diverse computational resources and user backgrounds, achieving efficient memory usage and processing speed while supporting multiple programming languages; and (3) \textbf{an active open-source community} fostering reproducible healthcare AI research through collaborative development with over 50+ examples and tutorials that lower domain expertise barriers.

\section{Related Works} \label{sec:related works}
\begin{table*}[h]
\centering
\caption{\textbf{Comparison of Healthcare AI Frameworks.}
``Dynamically Scalable to Consumer Hardware'' refers specifically to datasets
that exceed the memory capacity of standard machines; frameworks without this
support may still process large datasets given sufficient RAM on a workstation.
Clinical imaging datasets often contain lightweight metadata files such that
reading from disk occurs by design, meaning the memory footprint is usually
quite small. In contrast, clinical EHR datasets contain massive tables that
surpass memory constraints on even sizable workstations, which require further
engineering practices. For examples of unified multimodal data support,
interpretability, and uncertainty quantification, please see
Appendix~\ref{appdx: interp and multimodality}.}
\label{tab:framework_comparison}
\renewcommand{\arraystretch}{1.3}
\begin{tabular}{@{}lccccc@{}}
\toprule
\textbf{Feature} & \textbf{PyHealth 1.16} & \textbf{MEDS} & \textbf{Zensols} & \textbf{MonAI} & \textbf{PyHealth 2.0} \\
\midrule
Reproducibility Focused & \textcolor{ForestGreen}{\checkmark} & \textcolor{ForestGreen}{\checkmark} & \textcolor{ForestGreen}{\checkmark} & \textcolor{ForestGreen}{\checkmark} & \textcolor{ForestGreen}{\checkmark} \\
Biosignals & \textcolor{ForestGreen}{\checkmark} & \textcolor{red}{\ding{55}} & \textcolor{red}{\ding{55}} & \textcolor{red}{\ding{55}} & \textcolor{ForestGreen}{\checkmark} \\
EHR & \textcolor{ForestGreen}{\checkmark} & \textcolor{ForestGreen}{\checkmark} & \textcolor{red}{\ding{55}} & \textcolor{red}{\ding{55}} & \textcolor{ForestGreen}{\checkmark} \\
Clinical Text & \textcolor{red}{\ding{55}} & \textcolor{red}{\ding{55}} & \textcolor{ForestGreen}{\checkmark} & \textcolor{ForestGreen}{\checkmark} & \textcolor{ForestGreen}{\checkmark} \\
Clinical Imaging & \textcolor{red}{\ding{55}} & \textcolor{red}{\ding{55}} & \textcolor{red}{\ding{55}} & \textcolor{ForestGreen}{\checkmark} & \textcolor{ForestGreen}{\checkmark} \\
Unified Multimodal Data Support & \textcolor{red}{\ding{55}} & \textcolor{red}{\ding{55}} & \textcolor{red}{\ding{55}} & \textcolor{red}{\ding{55}} & \textcolor{ForestGreen}{\checkmark} \\
Large Selection of Models & \textcolor{ForestGreen}{\checkmark} & \textcolor{red}{\ding{55}} & \textcolor{red}{\ding{55}} & \textcolor{ForestGreen}{\checkmark} & \textcolor{ForestGreen}{\checkmark} \\
Interpretability & \textcolor{red}{\ding{55}} & \textcolor{red}{\ding{55}} & \textcolor{red}{\ding{55}} & \textcolor{ForestGreen}{\checkmark} & \textcolor{ForestGreen}{\checkmark} \\
Uncertainty Quantification & \textcolor{ForestGreen}{\checkmark} & \textcolor{red}{\ding{55}} & \textcolor{red}{\ding{55}} & \textcolor{red}{\ding{55}} & \textcolor{ForestGreen}{\checkmark} \\
Programming API Only & \textcolor{ForestGreen}{\checkmark} & \textcolor{red}{\ding{55}} & \textcolor{red}{\ding{55}} & \textcolor{ForestGreen}{\checkmark} & \textcolor{ForestGreen}{\checkmark} \\
End to End Pipelines & \textcolor{ForestGreen}{\checkmark} & \textcolor{red}{\ding{55}} & \textcolor{red}{\ding{55}} & \textcolor{ForestGreen}{\checkmark} & \textcolor{ForestGreen}{\checkmark} \\
Dynamically Scalable to Consumer Hardware & \textcolor{red}{\ding{55}} & \textcolor{ForestGreen}{\checkmark} & \textcolor{red}{\ding{55}} & \textcolor{red}{\ding{55}}\textsuperscript{*} & \textcolor{ForestGreen}{\checkmark} \\
\bottomrule
\end{tabular}
\end{table*}

PyHealth is not the only healthcare AI toolkit available for wider use. PyHealth 2.0's design draws inspiration from the broader healthcare AI reproducibility community, and we acknowledge these contributions while highlighting PyHealth's distinct design philosophy and interoperability with existing frameworks. We provide an overview of the differences of PyHealth 2.0 with respect to many other health AI frameworks in Table \ref{tab:framework_comparison}.

\textbf{Other Health AI Frameworks.} Several specialized frameworks address different aspects of healthcare AI development. The \textbf{MEDS ecosystem} \citep{mcdermott2025meds} provides a suite of modular tools for developing and benchmarking models on longitudinal EHR data, operating as a network of interoperable packages linked by a common event stream data standard. Notable components include ACES for automatic cohort extraction \citep{xu2024aces} and MEDS-tab for automated baseline training of tabular models \citep{oufattole2024meds_tab}. \textbf{MONAI} \citep{cardoso2022monai} specializes in clinical imaging, offering tools for segmentation, classification, and generative modeling of medical images. \textbf{Zensols} \cite{landes2023deepzensols} focuses on rapid reproduction of traditional clinical NLP models for text-based clinical problems.

PyHealth distinguishes itself through its flexibility and scope. By adopting a highly flexible event stream format \cite{arnrich2024medical_meds_standard}, PyHealth maintains fundamental interoperability with these frameworks while offering unique advantages. Unlike MEDS's focus on longitudinal structured EHR data \citep{mcdermott2025meds}, MONAI's specialization in imaging \citep{cardoso2022monai}, or Zensols's emphasis on text \cite{landes2023deepzensols}, PyHealth's lack of datatype assumptions enables seamless integration of any combination of signals, clinical notes, lab events, structured medical codes, and images. Additionally, PyHealth requires only Python knowledge, eliminating the need to learn separate definition schemas such as those required by tools like ACES \citep{xu2024aces}.

\textbf{Data Standards.} PyHealth makes no assumptions on data standards, and its flexible internal data structures support integration with multiple established clinical data standards. Common data models like OMOP \citep{reinecke2021usage_omop}, FHIR \citep{bender2013hl7_fhir}, i2b2 \citep{murphy2010serving_i2b2}, and PCORNet \citep{forrest2021pcornet} exist to standardize clinical vocabularies and longitudinal data storage across different institutions, making it easier to transfer models and perform data analysis across different institutions. PyHealth currently supports the OMOP standard directly while other data standards can be adapted without difficulty.

\textbf{Benchmarks.} Other software focus on benchmarking clinical models. YAIB \citep{vandewater2024icubenchmarkflexiblemulticenter_YAIB}, EHR-PT \citep{mcdermott2021comprehensive_ehr_pt}, and a Multitask Clinical Benchmark \citep{harutyunyan2019multitask_clinical_bench} were constructed to standardize model performance benchmarking across popular clinical datasets. We note that PyHealth models are compatible with the majority of these benchmarks. One crucial differentiator here is that the PyHealth API enables easy modifications to existing defined benchmarks, making it flexible for a variety of clinical needs. As a key consequence, PyHealth can also be leveraged to quickly reproduce each of these benchmarks. Furthermore, PyHealth's evaluation goes beyond performance benchmarking, supporting model interpretability and uncertainty quantification, post-hoc analyses that enable a deeper look within a clinical model. 

\textbf{What Makes PyHealth Different.} PyHealth's design philosophy centers on \textit{comprehensiveness}: providing a unified framework that integrates diverse data modalities, model architectures, and analytical frameworks within a single cohesive system. This comprehensive approach offers several key advantages. First, researchers can work with multiple data types—structured EHR events, clinical notes, medical images, time series signals—without switching between specialized frameworks or managing complex dependency chains. Second, the unified API eliminates the overhead of learning multiple tools or reconciling different data formats across frameworks, allowing researchers to focus on model development rather than infrastructure. Third, by maintaining compatibility with established standards (OMOP \citep{reinecke2021usage_omop}, FHIR \citep{bender2013hl7_fhir}) and interoperability with existing tools (MEDS \citep{mcdermott2025meds}, MONAI \citep{cardoso2022monai}), PyHealth enables integration into existing workflows while providing the convenience of a single, actively-maintained package.

As a comprehensive yet modular framework, PyHealth delivers integrated benefits spanning the complete modeling pipeline: optimized data processing, streamlined baseline reproduction, model interpretability tools, and uncertainty quantification methods. The open-source PyHealth community further strengthens this ecosystem through active peer-reviewed research that validates each contribution to the software package. Ultimately, PyHealth 2.0 provides a comprehensive suite of APIs that enables users to rapidly develop reproducible clinical models directly in their coding environments, facilitating adoption in both research and production settings. 
\section{PyHealth 2.0} \label{sec:methodology}

\begin{figure*}[p]
    \centering
    \includegraphics[width=0.95\textwidth]{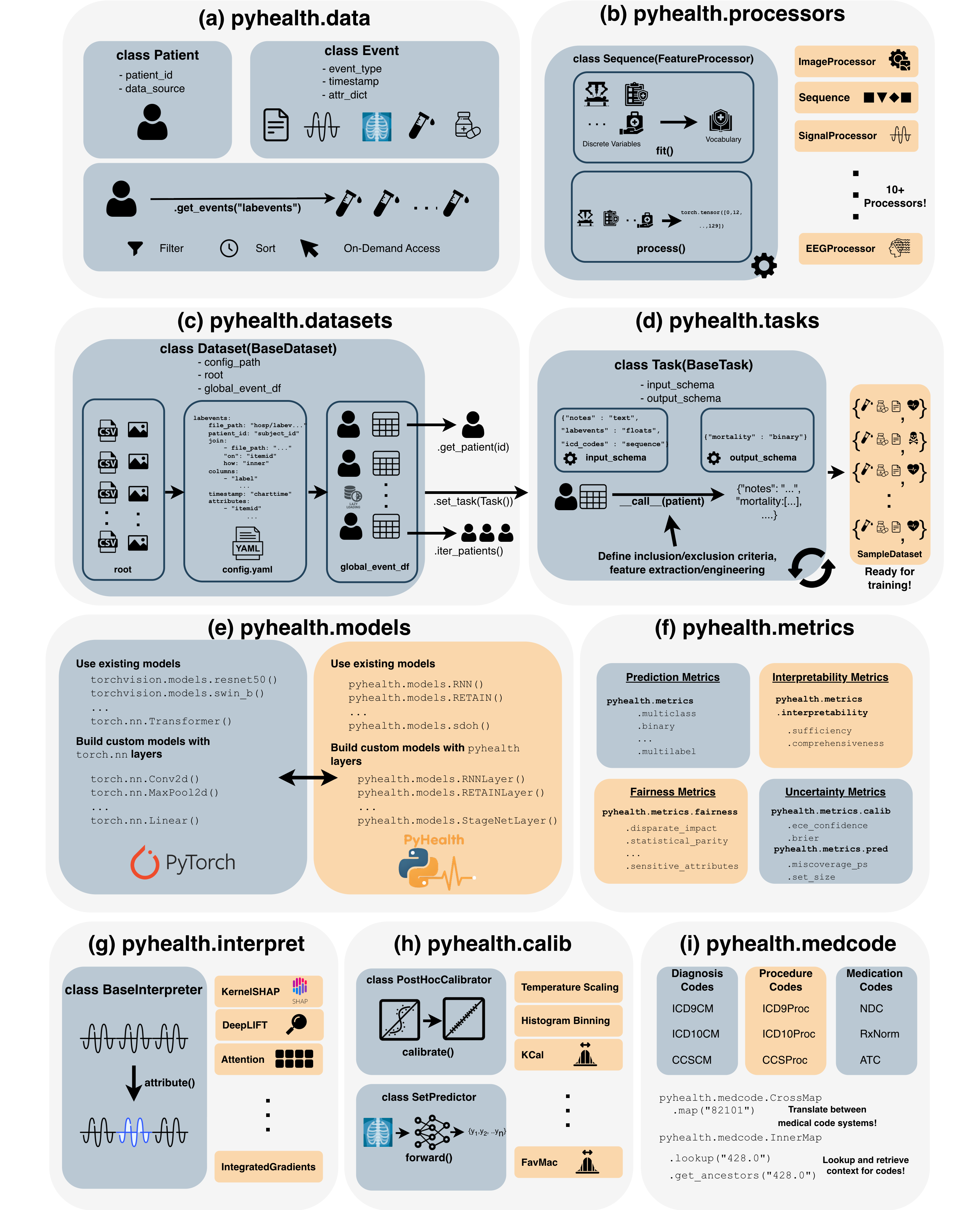}
    \caption{\textbf{PyHealth Overview.} PyHealth provides a comprehensive end-to-end pipeline for healthcare AI development spanning all aspects of clinical modeling. Starting from (a) standardized data structures (pyhealth.data) and (b) flexible data transformations (pyhealth.processors), we accommodate diverse clinical datasets through (c) memory-efficient lazy-loaded dataset loading (pyhealth.datasets) and (d) optimized task-specific ML processing for creating trainable formats (pyhealth.tasks). The framework supports (e) baseline model benchmarking (pyhealth.models) and (f) comprehensive evaluation metrics for fairness, uncertainty quantification, and interpretability—all critical for clinical deployment (pyhealth.metrics). For post-training utility, we provide (g) model interpretation tools (pyhealth.interpret), (h) uncertainty quantification via calibration and conformal prediction (pyhealth.calib), and (i) medical code translation and lookup across hospital coding standards (pyhealth.medcode).}
    \label{fig:pyhealth_framework}
\end{figure*}

PyHealth 2.0 currently consists of 9 major modules for comprehensive clinical model deployment, spanning from raw data processing to complex model evaluation, as shown in Figure \ref{fig:pyhealth_framework}. The PyHealth modules can be organized within two separate aspects of deep learning: model training and model evaluation. Each module has a set of base APIs that can be extended to support custom workflows.

\textbf{Model Training.} At its core, model training requires processing raw data into a trainable format (i.e torch tensors). To do so, PyHealth supports the following modules to take a raw dataset and perform training in 7 lines of code as shown in Figure \ref{fig:pyhealth_benefits}.

\textbf{pyhealth.data} provides a flexible 2-layer hierarchical data structure. Like the MEDs schema \citep{arnrich2024medical_meds_standard}, everything is organized around patients, where each Patient contains a set of Event objects. Unlike MEDs, PyHealth makes no assumptions about dates, data types, or formats, enabling support for diverse clinical datasets including signals, images, structured EHR, and other modalities. For datasets without patient\_id, each sample or row is treated as a Patient with a single event.

\textbf{pyhealth.processors} enables rapid transformation between data types through normalization, tokenization, and other clinical model requirements. Most processors support direct translation of continuous or discrete variables into torch tensors for training.

\textbf{pyhealth.datasets} introduces an optimized lazy-loading solution in PyHealth 2.0 for both dataset exploration and task construction. By loading each patient into memory only when needed, entire dataset tables load nearly instantaneously while ML task processing dynamically adapts to different memory constraints. As task processing can be expensive across millions of events, our dataset class supports parallel task processing, enabling massive speedups as shown in Figure \ref{fig:pyhealth_performance}.

\textbf{pyhealth.tasks} provides a simple, readable interface for defining diverse machine learning tasks through three components: (1) an input schema specifying expected inputs and pyhealth.processors for sample transformation, (2) an output schema defining the objective (generation, classification, etc.), and (3) a call function describing the sample construction process (feature inclusion, engineering, transformations, etc.). This design ensures that practitioners can immediately understand the inputs, outputs, and feature inclusion logic of any clinical ML task simply by reading its implementation. Once defined, the set\_task() function performs this task transformation in parallel, generating a SampleDataset object that inherits from a PyTorch Lightning streaming dataset class to ensure memory-efficient data iteration \citep{litdata2023}.

\textbf{pyhealth.models} are custom PyTorch \citep{paszke2019pytorch} models, making them interchangeable with any PyTorch model while providing the modularity needed to transition between frameworks. This design enables direct compatibility with distributed training frameworks such as PyTorch Lightning \citep{falcon2019pytorch_lightning}.

\textbf{Model Evaluation.} PyHealth offers a variety of ways of evaluating a model from interpretability, predictive performance, uncertainty quantification, to even translating medical codes to better understand a model's inputs. Below, we discuss each of the modules that enable comprehensive evaluation of a deep clinical model.

\textbf{pyhealth.metrics} offers comprehensive model evaluation beyond standard classification metrics for binary, multiclass, and multilabel settings. The module includes fairness metrics, interpretability metrics, and uncertainty quantification metrics to assess how fair, interpretable, and well-calibrated models are—critical considerations for clinical deployment.

\textbf{pyhealth.interpret} provides qualitative feature attribution analysis to complement quantitative evaluation. Interpreting deep learning models typically requires direct layer access (attention weights, convolution feature maps, intermediate gradients), specific input formatting, or navigation of incompatible dependencies across libraries like SHAP \citep{lundberg2017unifiedapproachinterpretingmodel_shap} and Captum \citep{kokhlikyan2020captum}. To address this, we directly implement popular interpretability approaches—including Attention-Grad \citep{chefer2021genericattentionmodelexplainabilityinterpreting_attngrad}, GIM \citep{edin2025gimimprovedinterpretabilitylarge_GIM}, DeepLift \citep{shrikumar2017deeplift}, and SHAP \citep{lundberg2017unifiedapproachinterpretingmodel_shap}—for seamless integration with clinical models.

\textbf{pyhealth.calib} addresses uncertainty quantification, which is critical for deployment in high-risk clinical settings. The module supports multiple techniques including model calibration \citep{wang2023modelcalibration, guo2017calibration} and conformal prediction \citep{angelopoulos2023conformal}.

\textbf{pyhealth.medcode} handles medical coding standard translations. Clinical data uses various standards, most commonly diagnosis codes like International Classification of Diseases (ICD) \citep{cartwright2013icd} and Clinical Classification Software (CCS) \citep{wei2017evaluating_ccsm}, or drug codes like ATC \citep{miller1995new_atc}, RXNorm \citep{nelson2011normalized_rxnorm}, and NDC \citep{simonaitis2009using_ndc}. PyHealth supports direct translation between these ontologies plus definition and ancestor lookups for contextualizing code inputs and predictions.

We further describe other implementation details in Appendix section \ref{appdx : dataloading and caching details} and how users can extend PyHealth for their own use cases in Appendix sections \ref{appdx: datasets}, \ref{appdx: tasks}, \ref{appdx: models}.
\section{Results} \label{sec:results}
\textbf{PyHealth 2.0 offers substantial improvements over PyHealth 1.16 in functionality and accessibility.}
To drive adoption, a software repository must provide compelling capabilities that justify its use. As discussed in Table \ref{tab:framework_comparison}, PyHealth 2.0 introduces several key advantages over existing healthcare AI frameworks. Unlike MEDS \citep{mcdermott2025meds}, MonAI \citep{cardoso2022monai}, and PyHealth 1.16 \citep{yang2023pyhealth}, it supports true multimodal data integration—combining images, biosignals, structured codes, and clinical notes within a single dataloader. This enables researchers to explore richer feature combinations in their clinical pipelines.

Beyond data handling, PyHealth 2.0 provides an expanding model library and comprehensive post-hoc deployment tools including interpretability and uncertainty quantification. The updated toolkit scales dynamically across different hardware configurations. Most significantly, PyHealth 2.0 addresses the memory management issues identified by \citet{steinberg2024meds_reader} in PyHealth 1.16, enabling users to train clinical predictive models on consumer-grade hardware where memory is limited.

\textbf{Performance Benchmarking Setup.} To demonstrate the performance benefits of PyHealth 2.0, we benchmark on MIMIC-IV version 2.2 \citep{johnson2023mimic4} using an AMD EPYC 7513 32-core processor workstation with 1TB of RAM. This dataset contains 315,460 patients, 454,324 admissions, 5,006,884 diagnosis codes, 704,124 procedure codes, 124,342,638 lab events, and 669,186 drug codes. We evaluate three tasks: mortality prediction (using lab events and ICD codes), drug recommendation, and length-of-stay prediction. Each task requires joining multiple tables and iteratively extracting patients who have all necessary events. Of these, mortality prediction is the most expensive in both time and memory due to its use of lab events.

\begin{figure*}[h!]
    \centering
    \includegraphics[width=1.0\textwidth]{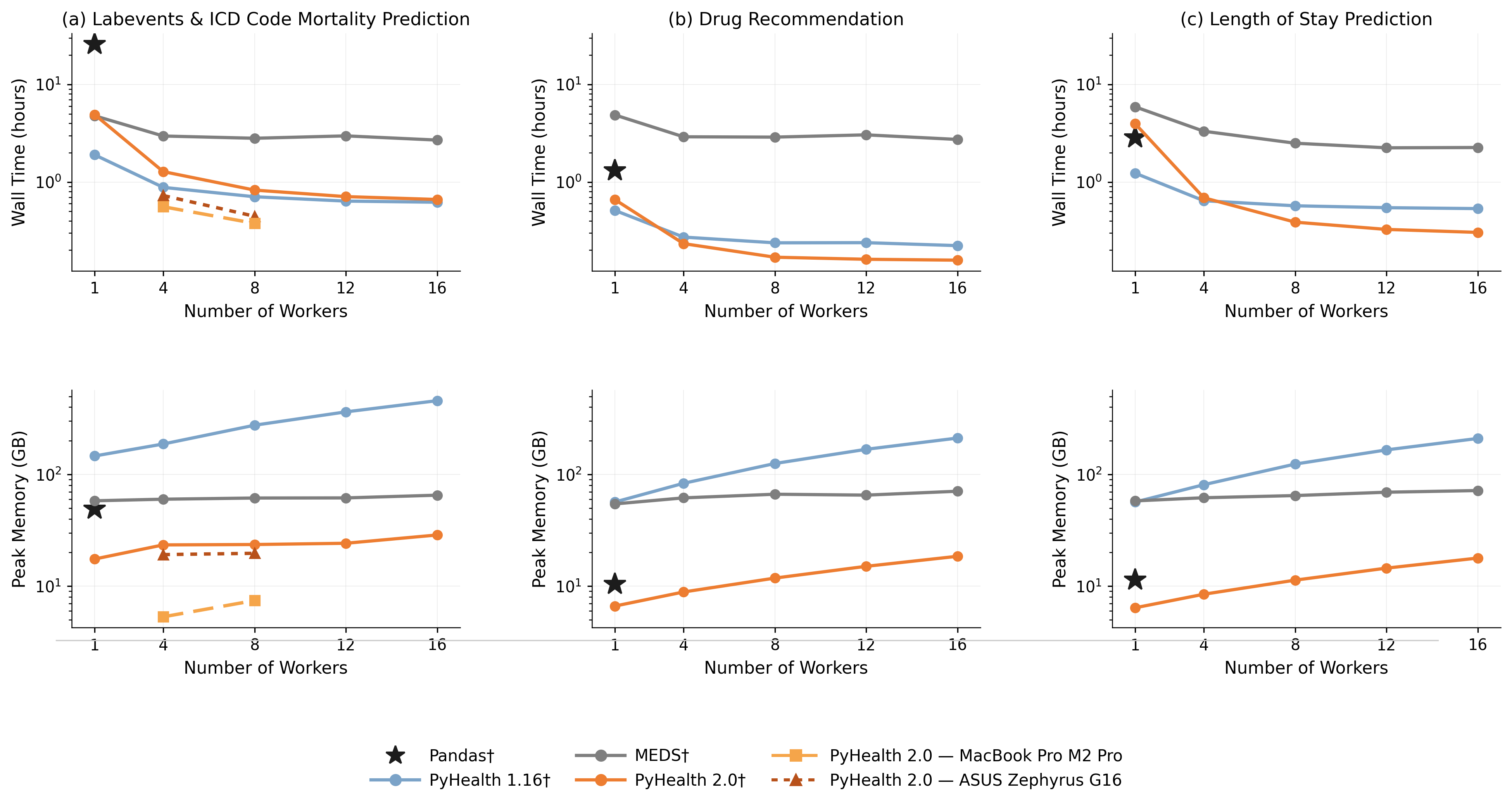}
    \caption{\textbf{PyHealth 2.0 delivers efficient scaling across machines.} We benchmark
    PyHealth 2.0 against PyHealth 1.16, MEDS, and a naive Pandas baseline --- each
    processing raw MIMIC-IV \citep{johnson2023mimic4} \texttt{.csv} files through to
    trainable tensor formats --- across three tasks: (a) labevents-based mortality
    prediction, (b) drug recommendation, and (c) length of stay prediction. All baselines
    and the Large Workstation (WS, NAS, DDR4) runs use a server-class machine; panel~(a)
    additionally reports PyHealth~2.0 results on a MacBook Pro M2~Pro (16\,GB) and an
    ASUS Zephyrus G16 (32\,GB DDR5). PyHealth~2.0 achieves up to 39$\times$ speedup over
    naive Pandas and remains comparable to or faster than PyHealth~1.16 on wall time.
    Crucially, PyHealth~2.0 running on a consumer MacBook Pro M2~Pro consumes over
    35$\times$ less peak memory than PyHealth~1.16 on a server-class workstation,
    a platform that previously required dedicated server infrastructure now fits
    comfortably within a 16\,GB laptop memory budget. This is enabled by multi-worker
    sharding via Dask~\citep{rocklin2015dask} and Polars~\citep{nahrstedt2024empirical_polars},
    which keep memory consumption nearly constant as worker count scales. Exact numbers
    are provided in Appendix~\ref{appdx: pyhealth table perf}.
    $\dagger$\,Pandas, PyHealth~1.16, and MEDS are benchmarked on the WS only.}
    \label{fig:pyhealth_performance}
\end{figure*}

\textbf{PyHealth 2.0 scales efficiently from laptops to large compute clusters.}
Addressing the reproducibility crisis requires lowering barriers to entry for training clinical AI models. PyHealth 2.0 achieves this through substantial performance improvements: task processing scales more efficiently across multiple workers than PyHealth 1.16, using significantly less memory while better utilizing available CPU cores for data transformation. Critically, our benchmark demonstrates end-to-end efficiency—the system not only extracts patient cohorts but directly translates and caches data into trainable tensor formats, making subsequent task reuse instantaneous after initial processing. In Figure \ref{fig:pyhealth_performance}, PyHealth 2.0's memory usage remains relatively constant across worker counts, and is consistently faster at ML task processing than the naive Pandas approach and its previous PyHealth 1.16 version across the majority of worker counts across all three tasks. We observe that the more modular MEDS counterpart, which includes the use of two frameworks MEDS\_ETL and MEDS\_reader \citep{steinberg2024meds_reader}, while substantially more memory-efficient than PyHealth 1.16, trades this memory efficiency for less multi-core performance.

\textbf{Consumer hardware outperforms server infrastructure with PyHealth 2.0.}
A striking finding from our laptop task processing benchmarks is that PyHealth 2.0 running on a 16\,GB MacBook Pro M2 Pro is \emph{faster} on the labevents mortality task than our 32-core server workstation. This arises from three hardware factors: (1) the workstation accesses data over a NAS, introducing network I/O bottlenecks, whereas laptops read from local NVMe SSDs; (2) laptops offer higher memory bandwidth (unified LPDDR5/DDR5 vs.\ DDR4); and (3) PyHealth 2.0's sequential streaming workload is bottlenecked by single-core throughput, where modern consumer hardware excels. At 8 workers, the MacBook Pro completes the full labevents task in 1,357\,s—faster than PyHealth 1.16 at 16 workers (2,235\,s) on the server—while consuming a fraction of the memory. Researchers without dedicated server infrastructure can therefore achieve competitive or superior throughput on a standard laptop, substantially broadening access to clinical AI development.

\begin{table*}[h!]
\centering
\caption{\textbf{PyHealth 2.0 standardizes clinical pipeline deployment with
fewer lines of code.} Compared to Pandas, PyHealth 1.16, and the MEDS
ecosystem (MEDS\_etl \citep{mcdermott2025meds} and MEDS\_Reader
\citep{steinberg2024meds_reader}), PyHealth 2.0 achieves the fewest lines for
patient data exploration and a uniform 7-line initialization across all tasks.
PyHealth 1.16 yields slightly fewer lines on ML tasks due to its functional
(non-OOP) style; by contrast, PyHealth 2.0's OOP design assigns each task
variant a \textbf{unique identifier} (the primary source of additional
lines), which is essential for reproducibility and extensibility as tasks and
modalities scale.
\textbf{Bold}: lowest count per column;
\textcolor{green}{$\downarrow$}: improvement over Pandas and MEDS.
*Pre-implemented task logic accounts for most ML code savings; custom tasks
require only one additional function
(Figure~\ref{fig:pyhealth_patient_data_task_construction}).}
\label{tab:loc_comparison}
\small
\begin{tabular}{@{}lcccc@{}}
\toprule
Method & Patient Exploration & Mortality Prediction\textasteriskcentered{} & Length of Stay\textasteriskcentered{} & Drug Recommendation\textasteriskcentered{} \\
\midrule
Pandas               & 16           & 51           & 22           & 24           \\
PyHealth 1.16        & 14           & \textbf{27}  & \textbf{14}  & \textbf{16}  \\
MEDS ETL + MEDS\_Reader & 12        & 43           & 38           & 39           \\
PyHealth 2.0         & \textbf{10} \textcolor{green}{$\downarrow$} & 34 \textcolor{green}{$\downarrow$} & 18 \textcolor{green}{$\downarrow$} & 23 \textcolor{green}{$\downarrow$} \\
\bottomrule
\end{tabular}
\end{table*}

\begin{figure*}[h!]
    \centering
    \includegraphics[width=1.0\textwidth]{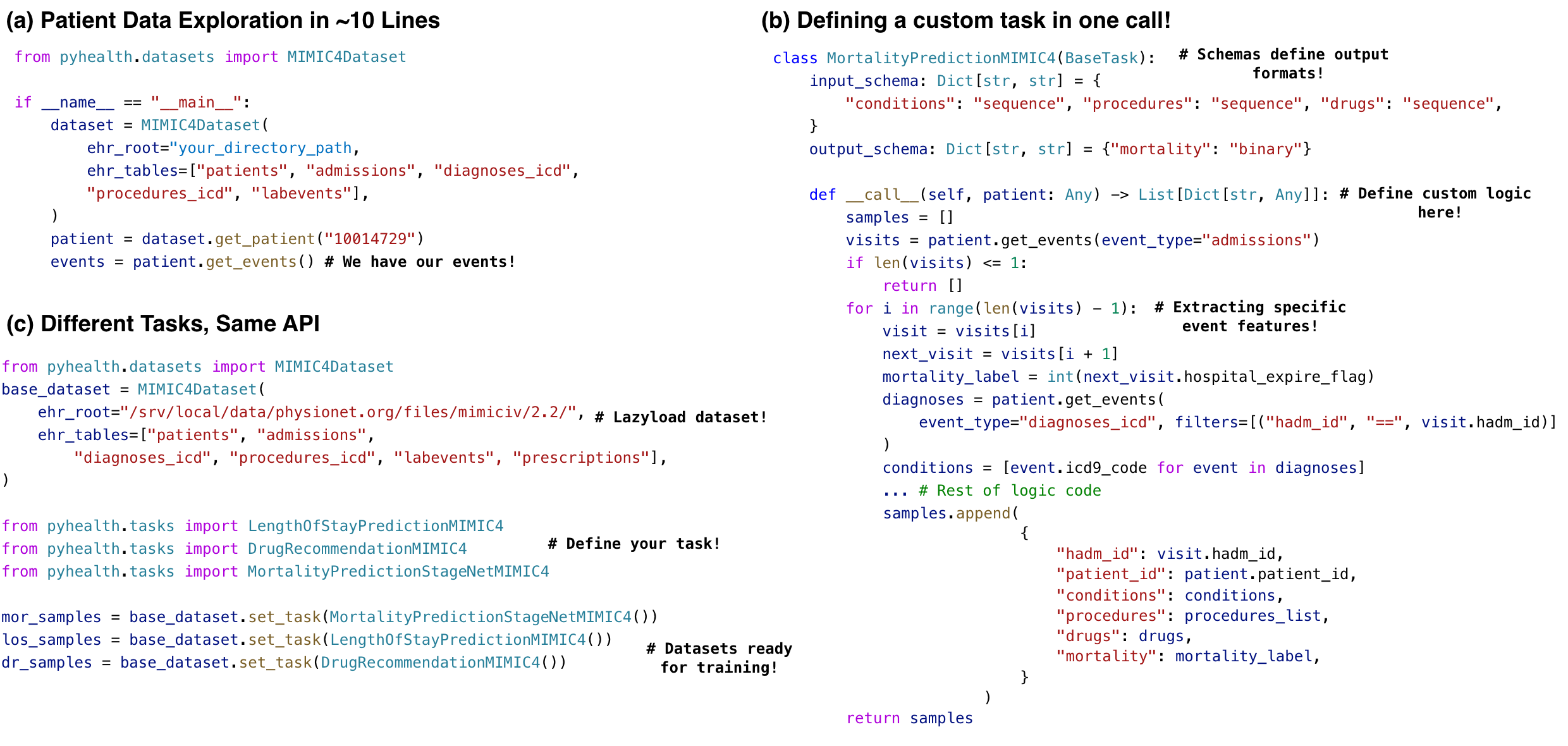}
    \caption{\textbf{Defining your own PyHealth tasks.} (a) Exploring patient data is key to defining your own custom task. (b) Once a task class is defined, PyHealth 2.0's optimized backend can handle the rest of the efficient parallel data processing. The key approach here is that for any new task, only a single class call has to be defined, following the (c) same pipeline for generating each task here.  }
    \label{fig:pyhealth_patient_data_task_construction}
\end{figure*}

\textbf{PyHealth 2.0 offers streamlined patient data exploration.} Exploring patient data in large disaggregated EHR datasets requires aggregating event data from multiple sources. This involves joining multiple tables, querying a specific patient, and loading each relevant event into a single data structure. Our updated PyHealth API has further streamlined patient data exploration, reducing the required code to aggregate all patient events for a given patient in MIMIC-IV \citep{johnson2023mimic4} from 14 lines in PyHealth 1.16 \citep{yang2023pyhealth} to only 10 in our 2.0 release, as shown in Table \ref{tab:loc_comparison} and Figure \ref{fig:pyhealth_patient_data_task_construction} (a). With a single .get\_events() call, users can explore all patient events from the specified tables lazy-loaded by the dataset object.

\textbf{Defining a new reproducible ML task is straightforward with PyHealth.} A key goal in improving healthcare AI accessibility is reducing the complexity of defining tasks for downstream training. In Table \ref{tab:loc_comparison}, the number of lines of code to use an already-implemented task remains constant regardless of task type. To define a new task, only a single function call is required to leverage all optimized backend processing, as shown in Figure \ref{fig:pyhealth_patient_data_task_construction} (b). By standardizing how each task is called within the rest of the pipeline (Figure \ref{fig:pyhealth_patient_data_task_construction} (c)), users can rapidly prototype and experiment with different features for modeling purposes. In Appendix \ref{appdx: interp and multimodality}, we demonstrate how users can quickly define a mortality prediction task that incorporates all modalities within MIMIC-IV \citep{johnson2023mimic4}, including clinical notes, X-rays, lab events, and structured EHR codes. In  Appendix \ref{app:benchmarks}, we  show benchmark results across various healthcare ML tasks, uncertainty quantification methods, and interpretability approaches.

\section{Discussion}
PyHealth 2.0 addresses fundamental barriers in clinical AI through unified multimodal capabilities, accessibility-focused design, and reproducible research infrastructure. These advances establish a foundation for tackling long-standing challenges in clinical AI deployment and reproducibility.

\textbf{Moving accessibility beyond Python-based communities.} Much of the bioinformatics community conducts research in R \citep{staples2023expansion_R}, limiting the ability to replicate and advance AI models on genomics \citep{akalin2020computational_R_genomics} and other modalities crucial to patient outcomes. To bridge this gap, the PyHealth community has introduced RHealth \citep{RHealth2025}, which brings PyHealth's core functionalities to R. As PyHealth develops, its design principles will extend to communities beyond Python-based machine learning.

\textbf{Future of Agentic Systems and PyHealth.} As coding agents rapidly improve \citep{yang2024if_coding_agents,wang2025agents_swe}, PyHealth's unified API provides an ideal substrate for automated clinical AI pipelines. Its modality-agnostic design—making minimal assumptions about data types or architectures—enables agents to automate more of the clinical deployment workflow. While data privacy prevents exact replication across healthcare systems, well-documented code serves as a practical recipe for institutional adaptation. PyHealth thus functions as a centralized repository of recipes for agentic systems.

\textbf{PyHealth Roadmap.} PyHealth development remains actively ongoing with many planned features. While the number of models, datasets, and tasks continues to expand as the community grows, several features orthogonal to ML pipelines are under development. PyHealth 2.0 currently supports various interpretability and uncertainty quantification methods as shown in Appendix Figure \ref{fig:cxr_example}, though current implementations have several limitations.

First, many interpretability approaches assume access to specific PyHealth model attributes, such as embedding layers or hooks enabling gradient logging. These assumptions limit model compatibility with the interpretability module, currently restricting support to tabular time-series clinical predictive models and image-based models. Support for interpreting language models and other modalities is a major future priority as the field of mechanistic interpretability develops \citep{rai2024practical_review_survey_mechinterp, fiotto2024nnsight}.

Second, many uncertainty quantification approaches such as conformal prediction \citep{angelopoulos2023conformal,papadopoulos2007conformal} fail on clinical tasks because patient distribution shifts are common. While approaches like conformal prediction under covariate shift exist within the repository, their practical utility remains limited on various clinical tasks. As a key priority, PyHealth intends to explore personalized conformal prediction approaches to make the uncertainty quantification module more useful where distribution assumptions are unlikely to hold.

Finally, while PyHealth directly supports mixing and matching any modality as shown in Appendix Figure \ref{fig:multimodal data loading}, its support for multimodal models capable of handling heterogeneous features remains highly limited. Though capable of extracting embeddings from HuggingFace \citep{jain2022huggingface} models and various provided unimodal PyTorch models, PyHealth does not readily provide recipes for off-the-shelf multimodal clinical predictions, especially under cases of missing modalities—highly prevalent in patient data \citep{wu2024deep_multimodal_missing}. Addressing this limitation is crucial for improving software usability and adoption.

\textbf{Limitations.} Beyond the modality, interpretability, and uncertainty quantification gaps outlined in our roadmap, two further limitations warrant attention. First, our processing and efficiency benchmarks are sensitive to hardware and operating system: storage medium, memory bandwidth, and core count each meaningfully affect wall-clock time and peak memory, so reported gains may vary across deployments. Second, PyHealth 2.0 deliberately scopes itself to data processing and modeling and does not provide data governance or de-identification tooling; users remain responsible for regulatory compliance (e.g., HIPAA, GDPR) and the ethical handling of patient data. Component maturity is similarly uneven—data processing, task schemas, and evaluation are extensively tested, and EHR and EEG modeling are well-supported, while clinical imaging, text, and genomics currently provide only basic baselines—and we will continue to document this clearly as the framework matures.

\textbf{Concluding Statements.} We present PyHealth 2.0, a vastly improved comprehensive clinical deep learning toolkit. This software enables reproducible development of clinical pipelines in a fast and accessible manner. Our hope is that by contributing this repository to the wider community, we can incentivize sharing of reproducible models and, more importantly, operable code in the AI for healthcare community.

\section*{Impact Statement}
PyHealth aims to reduce the complexity of developing clinical AI models. In doing so, PyHealth has reduced computational requirements, improving the accessibility of building clinical AI models. Our goal is to democratize the ability to work in clinical AI.




\nocite{langley00}

\bibliography{sections/bib}

@inproceedings{langley00,
 author    = {P. Langley},
 title     = {Crafting Papers on Machine Learning},
 year      = {2000},
 pages     = {1207--1216},
 editor    = {Pat Langley},
 booktitle     = {Proceedings of the 17th International Conference
              on Machine Learning (ICML 2000)},
 address   = {Stanford, CA},
 publisher = {Morgan Kaufmann}
}

@inproceedings{wang2024recent_survey_clinical_predictive_modeling,
  title={Recent advances in predictive modeling with electronic health records},
  author={Wang, Jiaqi and Luo, Junyu and Ye, Muchao and Wang, Xiaochen and Zhong, Yuan and Chang, Aofei and Huang, Guanjie and Yin, Ziyi and Xiao, Cao and Sun, Jimeng and others},
  booktitle={IJCAI: proceedings of the conference},
  volume={2024},
  pages={8272},
  year={2024}
}

@article{janiesch2021machine,
  title={Machine learning and deep learning},
  author={Janiesch, Christian and Zschech, Patrick and Heinrich, Kai},
  journal={Electronic markets},
  volume={31},
  number={3},
  pages={685--695},
  year={2021},
  publisher={Springer}
}

@inproceedings{shinde2018review_dl_apps,
  title={A review of machine learning and deep learning applications},
  author={Shinde, Pramila P and Shah, Seema},
  booktitle={2018 Fourth international conference on computing communication control and automation (ICCUBEA)},
  pages={1--6},
  year={2018},
  organization={IEEE}
}

@misc{mcdermott2019reproducibilitymachinelearninghealth,
      title={Reproducibility in Machine Learning for Health}, 
      author={Matthew B. A. McDermott and Shirly Wang and Nikki Marinsek and Rajesh Ranganath and Marzyeh Ghassemi and Luca Foschini},
      year={2019},
      eprint={1907.01463},
      archivePrefix={arXiv},
      primaryClass={cs.LG},
      url={https://arxiv.org/abs/1907.01463}, 
}

@article{mcdermott2021reproducibility,
  title={Reproducibility in machine learning for health research: Still a ways to go},
  author={McDermott, Matthew BA and Wang, Shirly and Marinsek, Nikki and Ranganath, Rajesh and Foschini, Luca and Ghassemi, Marzyeh},
  journal={Science translational medicine},
  volume={13},
  number={586},
  pages={eabb1655},
  year={2021},
  publisher={American Association for the Advancement of Science}
}

@article{beam2020challenges_reproducibility,
  title={Challenges to the reproducibility of machine learning models in health care},
  author={Beam, Andrew L and Manrai, Arjun K and Ghassemi, Marzyeh},
  journal={Jama},
  volume={323},
  number={4},
  pages={305--306},
  year={2020},
  publisher={American Medical Association}
}

@misc{vandewater2024icubenchmarkflexiblemulticenter_YAIB,
      title={Yet Another ICU Benchmark: A Flexible Multi-Center Framework for Clinical ML}, 
      author={Robin van de Water and Hendrik Schmidt and Paul Elbers and Patrick Thoral and Bert Arnrich and Patrick Rockenschaub},
      year={2024},
      eprint={2306.05109},
      archivePrefix={arXiv},
      primaryClass={cs.LG},
      url={https://arxiv.org/abs/2306.05109}, 
}

@incollection{nissar2023bridging_gap_between_tech_and_medicine,
  title={Bridging the gap between technology and medicine: approaches of artificial intelligence in healthcare},
  author={Nissar, Iqra and Alam, Shahzad and Masood, Sarfaraz and Mir, Waseem Ahmad},
  booktitle={Machine learning and artificial intelligence in healthcare systems},
  pages={173--190},
  year={2023},
  publisher={CRC Press}
}

@InProceedings{pmlr-v68-johnson17a_reprod_mortality_pred,
  title = 	 {Reproducibility in critical care: a mortality prediction case study},
  author = 	 {Johnson, Alistair E. W. and Pollard, Tom J. and Mark, Roger G.},
  booktitle = 	 {Proceedings of the 2nd Machine Learning for Healthcare Conference},
  pages = 	 {361--376},
  year = 	 {2017},
  editor = 	 {Doshi-Velez, Finale and Fackler, Jim and Kale, David and Ranganath, Rajesh and Wallace, Byron and Wiens, Jenna},
  volume = 	 {68},
  series = 	 {Proceedings of Machine Learning Research},
  month = 	 {18--19 Aug},
  publisher =    {PMLR},
  url = 	 {https://proceedings.mlr.press/v68/johnson17a.html}
}

@article{boettiger2015introduction_docker_reproducibility,
  title={An introduction to Docker for reproducible research},
  author={Boettiger, Carl},
  journal={ACM SIGOPS Operating Systems Review},
  volume={49},
  number={1},
  pages={71--79},
  year={2015},
  publisher={ACM New York, NY, USA}
}

@article{acosta2022multimodal_biomed,
  title={Multimodal biomedical AI},
  author={Acosta, Juli{\'a}n N and Falcone, Guido J and Rajpurkar, Pranav and Topol, Eric J},
  journal={Nature medicine},
  volume={28},
  number={9},
  pages={1773--1784},
  year={2022},
  publisher={Nature Publishing Group US New York}
}

@article{johnson2023mimic4,
  title={MIMIC-IV, a freely accessible electronic health record dataset},
  author={Johnson, Alistair EW and Bulgarelli, Lucas and Shen, Lu and Gayles, Alvin and Shammout, Ayad and Horng, Steven and Pollard, Tom J and Hao, Sicheng and Moody, Benjamin and Gow, Brian and others},
  journal={Scientific data},
  volume={10},
  number={1},
  pages={1},
  year={2023},
  publisher={Nature Publishing Group UK London}
}

@inproceedings{mcdermott2025meds,
  title={Meds: Building models and tools in a reproducible health ai ecosystem},
  author={McDermott, Matthew BA and Xu, Justin and Bergamaschi, Teya S and Jeong, Hyewon and Lee, Simon A and Oufattole, Nassim and Rockenschaub, Patrick and Stankevi{\v{c}}i{\=u}t{\.e}, Kamil{\.e} and Steinberg, Ethan and Sun, Jimeng and others},
  booktitle={Proceedings of the 31st ACM SIGKDD Conference on Knowledge Discovery and Data Mining V. 2},
  pages={6243--6244},
  year={2025}
}

@article{xu2024aces,
  title={ACES: Automatic cohort extraction system for event-stream datasets},
  author={Xu, Justin and Gallifant, Jack and Johnson, Alistair EW and McDermott, Matthew},
  journal={arXiv preprint arXiv:2406.19653},
  year={2024}
}

@article{oufattole2024meds_tab,
  title={MEDS-Tab: Automated tabularization and baseline methods for MEDS datasets},
  author={Oufattole, Nassim and Bergamaschi, Teya and Kolo, Aleksia and Jeong, Hyewon and Gaggin, Hanna and Stultz, Collin M and McDermott, Matthew},
  journal={arXiv preprint arXiv:2411.00200},
  year={2024}
}

@article{cardoso2022monai,
  title={Monai: An open-source framework for deep learning in healthcare},
  author={Cardoso, M Jorge and Li, Wenqi and Brown, Richard and Ma, Nic and Kerfoot, Eric and Wang, Yiheng and Murrey, Benjamin and Myronenko, Andriy and Zhao, Can and Yang, Dong and others},
  journal={arXiv preprint arXiv:2211.02701},
  year={2022}
}

@article{reinecke2021usage_omop,
  title={The usage of OHDSI OMOP--a scoping review},
  author={Reinecke, Ines and Zoch, Mich{\'e}le and Reich, Christian and Sedlmayr, Martin and Bathelt, Franziska},
  journal={German Medical Data Sciences 2021: Digital Medicine: Recognize--Understand--Heal},
  pages={95--103},
  year={2021},
  publisher={IOS Press}
}

@article{pollard2018eicu,
  title={The eICU Collaborative Research Database, a freely available multi-center database for critical care research},
  author={Pollard, Tom J and Johnson, Alistair EW and Raffa, Jesse D and Celi, Leo A and Mark, Roger G and Badawi, Omar},
  journal={Scientific data},
  volume={5},
  number={1},
  pages={1--13},
  year={2018},
  publisher={Nature Publishing Group}
}

@inproceedings{bender2013hl7_fhir,
  title={HL7 FHIR: An Agile and RESTful approach to healthcare information exchange},
  author={Bender, Duane and Sartipi, Kamran},
  booktitle={Proceedings of the 26th IEEE international symposium on computer-based medical systems},
  pages={326--331},
  year={2013},
  organization={IEEE}
}

@article{murphy2010serving_i2b2,
  title={Serving the enterprise and beyond with informatics for integrating biology and the bedside (i2b2)},
  author={Murphy, Shawn N and Weber, Griffin and Mendis, Michael and Gainer, Vivian and Chueh, Henry C and Churchill, Susanne and Kohane, Isaac},
  journal={Journal of the American Medical Informatics Association},
  volume={17},
  number={2},
  pages={124--130},
  year={2010},
  publisher={BMJ Group}
}

@article{forrest2021pcornet,
  title={PCORnet{\textregistered} 2020: current state, accomplishments, and future directions},
  author={Forrest, Christopher B and McTigue, Kathleen M and Hernandez, Adrian F and Cohen, Lauren W and Cruz, Henry and Haynes, Kevin and Kaushal, Rainu and Kho, Abel N and Marsolo, Keith A and Nair, Vinit P and others},
  journal={Journal of Clinical Epidemiology},
  volume={129},
  pages={60--67},
  year={2021},
  publisher={Elsevier}
}

@article{harutyunyan2019multitask_clinical_bench,
  title={Multitask learning and benchmarking with clinical time series data},
  author={Harutyunyan, Hrayr and Khachatrian, Hrant and Kale, David C and Ver Steeg, Greg and Galstyan, Aram},
  journal={Scientific data},
  volume={6},
  number={1},
  pages={96},
  year={2019},
  publisher={Nature Publishing Group UK London}
}

@inproceedings{mcdermott2021comprehensive_ehr_pt,
  title={A comprehensive EHR timeseries pre-training benchmark},
  author={McDermott, Matthew and Nestor, Bret and Kim, Evan and Zhang, Wancong and Goldenberg, Anna and Szolovits, Peter and Ghassemi, Marzyeh},
  booktitle={Proceedings of the Conference on Health, Inference, and Learning},
  pages={257--278},
  year={2021}
}

@inproceedings{arnrich2024medical_meds_standard,
  title={Medical event data standard (MEDS): Facilitating machine learning for health},
  author={Arnrich, Bert and Choi, Edward and Fries, Jason Alan and McDermott, Matthew BA and Oh, Jungwoo and Pollard, Tom and Shah, Nigam and Steinberg, Ethan and Wornow, Michael and van de Water, Robin},
  booktitle={ICLR 2024 Workshop on Learning from Time Series For Health},
  pages={03--08},
  year={2024}
}

@article{paszke2019pytorch,
  title={Pytorch: An imperative style, high-performance deep learning library},
  author={Paszke, Adam and Gross, Sam and Massa, Francisco and Lerer, Adam and Bradbury, James and Chanan, Gregory and Killeen, Trevor and Lin, Zeming and Gimelshein, Natalia and Antiga, Luca and others},
  journal={Advances in neural information processing systems},
  volume={32},
  year={2019}
}

@article{falcon2019pytorch_lightning,
  title={Pytorch lightning},
  author={Falcon, William A},
  journal={GitHub},
  volume={3},
  year={2019}
}

@article{kokhlikyan2020captum,
  title={Captum: A unified and generic model interpretability library for pytorch},
  author={Kokhlikyan, Narine and Miglani, Vivek and Martin, Miguel and Wang, Edward and Alsallakh, Bilal and Reynolds, Jonathan and Melnikov, Alexander and Kliushkina, Natalia and Araya, Carlos and Yan, Siqi and others},
  journal={arXiv preprint arXiv:2009.07896},
  year={2020}
}

@misc{lundberg2017unifiedapproachinterpretingmodel_shap,
      title={A Unified Approach to Interpreting Model Predictions}, 
      author={Scott Lundberg and Su-In Lee},
      year={2017},
      eprint={1705.07874},
      archivePrefix={arXiv},
      primaryClass={cs.AI},
      url={https://arxiv.org/abs/1705.07874}, 
}

@misc{edin2025gimimprovedinterpretabilitylarge_GIM,
      title={GIM: Improved Interpretability for Large Language Models}, 
      author={Joakim Edin and Róbert Csordás and Tuukka Ruotsalo and Zhengxuan Wu and Maria Maistro and Casper L. Christensen and Jing Huang and Lars Maaløe},
      year={2025},
      eprint={2505.17630},
      archivePrefix={arXiv},
      primaryClass={cs.CL},
      url={https://arxiv.org/abs/2505.17630}, 
}

@misc{chefer2021genericattentionmodelexplainabilityinterpreting_attngrad,
      title={Generic Attention-model Explainability for Interpreting Bi-Modal and Encoder-Decoder Transformers}, 
      author={Hila Chefer and Shir Gur and Lior Wolf},
      year={2021},
      eprint={2103.15679},
      archivePrefix={arXiv},
      primaryClass={cs.CV},
      url={https://arxiv.org/abs/2103.15679}, 
}

@article{wang2023modelcalibration,
  title={Calibration in deep learning: A survey of the state-of-the-art},
  author={Wang, Cheng},
  journal={arXiv preprint arXiv:2308.01222},
  year={2023}
}

@inproceedings{guo2017calibration,
  title={On calibration of modern neural networks},
  author={Guo, Chuan and Pleiss, Geoff and Sun, Yu and Weinberger, Kilian Q},
  booktitle={International conference on machine learning},
  pages={1321--1330},
  year={2017},
  organization={PMLR}
}

@article{angelopoulos2023conformal,
  title={Conformal prediction: A gentle introduction},
  author={Angelopoulos, Anastasios N and Bates, Stephen and others},
  journal={Foundations and trends{\textregistered} in machine learning},
  volume={16},
  number={4},
  pages={494--591},
  year={2023},
  publisher={Now Publishers, Inc.}
}

@misc{cartwright2013icd,
  title={ICD-9-CM to ICD-10-CM codes: what? why? how?},
  author={Cartwright, Donna J},
  year={2013},
  publisher={Mary Ann Liebert, Inc. 140 Huguenot Street, 3rd Floor New Rochelle, NY 10801 USA}
}

@article{wei2017evaluating_ccsm,
  title={Evaluating phecodes, clinical classification software, and ICD-9-CM codes for phenome-wide association studies in the electronic health record},
  author={Wei, Wei-Qi and Bastarache, Lisa A and Carroll, Robert J and Marlo, Joy E and Osterman, Travis J and Gamazon, Eric R and Cox, Nancy J and Roden, Dan M and Denny, Joshua C},
  journal={PloS one},
  volume={12},
  number={7},
  pages={e0175508},
  year={2017},
  publisher={Public Library of Science San Francisco, CA USA}
}

@article{miller1995new_atc,
  title={A new drug classification for computer systems: the ATC extension code},
  author={Miller, GC and Britt, H},
  journal={International journal of bio-medical computing},
  volume={40},
  number={2},
  pages={121--124},
  year={1995},
  publisher={Elsevier}
}

@article{nelson2011normalized_rxnorm,
  title={Normalized names for clinical drugs: RxNorm at 6 years},
  author={Nelson, Stuart J and Zeng, Kelly and Kilbourne, John and Powell, Tammy and Moore, Robin},
  journal={Journal of the American Medical Informatics Association},
  volume={18},
  number={4},
  pages={441--448},
  year={2011},
  publisher={BMJ Group BMA House, Tavistock Square, London, WC1H 9JR}
}

@article{simonaitis2009using_ndc,
  title={Using National Drug Codes and drug knowledge bases to organize prescription records from multiple sources},
  author={Simonaitis, Linas and McDonald, Clement J},
  journal={American Journal of Health-System Pharmacy},
  volume={66},
  number={19},
  pages={1743--1753},
  year={2009},
  publisher={Oxford University Press}
}

@inproceedings{rocklin2015dask,
  title={Dask: Parallel computation with blocked algorithms and task scheduling.},
  author={Rocklin, Matthew and others},
  booktitle={SciPy},
  pages={126--132},
  year={2015}
}

@inproceedings{nahrstedt2024empirical_polars,
  title={An empirical study on the energy usage and performance of pandas and polars data analysis Python libraries},
  author={Nahrstedt, Felix and Karmouche, Mehdi and Bargie{\l}, Karolina and Banijamali, Pouyeh and Nalini Pradeep Kumar, Apoorva and Malavolta, Ivano},
  booktitle={Proceedings of the 28th international conference on evaluation and assessment in software engineering},
  pages={58--68},
  year={2024}
}

@misc{RHealth2025,
  author = {Ji Song and Zhixia Ren and Zhenbang Wu and John Wu and Chaoqi Yang and Jimeng Sun and Liantao Ma and Ewen M Harrison and Junyi Gao},
  title = {RHealth: A Deep Learning Toolkit for Healthcare Predictive Modeling},
  year = {2025},
  publisher = {GitHub},
  journal = {GitHub repository},
  howpublished = {\url{https://github.com/v1xerunt/RHealth}}
}

@article{staples2023expansion_R,
  title={Expansion and evolution of the R programming language},
  author={Staples, Timothy L},
  journal={Royal Society Open Science},
  volume={10},
  number={4},
  pages={221550},
  year={2023},
  publisher={The Royal Society}
}

@book{akalin2020computational_R_genomics,
  title={Computational genomics with R},
  author={Akalin, Altuna},
  year={2020},
  publisher={Chapman and Hall/CRC}
}

@article{yang2024if_coding_agents,
  title={If llm is the wizard, then code is the wand: A survey on how code empowers large language models to serve as intelligent agents},
  author={Yang, Ke and Liu, Jiateng and Wu, John and Yang, Chaoqi and Fung, Yi R and Li, Sha and Huang, Zixuan and Cao, Xu and Wang, Xingyao and Wang, Yiquan and others},
  journal={arXiv preprint arXiv:2401.00812},
  year={2024}
}

@article{wang2025agents_swe,
  title={Agents in software engineering: Survey, landscape, and vision},
  author={Wang, Yanlin and Zhong, Wanjun and Huang, Yanxian and Shi, Ensheng and Yang, Min and Chen, Jiachi and Li, Hui and Ma, Yuchi and Wang, Qianxiang and Zheng, Zibin},
  journal={Automated Software Engineering},
  volume={32},
  number={2},
  pages={70},
  year={2025},
  publisher={Springer}
}

@misc{litdata2023,
  author       = {Thomas Chaton and Lightning AI},
  title        = {LitData: Transform datasets at scale. Optimize datasets for fast AI model training.},
  year         = {2023},
  howpublished = {\url{https://github.com/Lightning-AI/litdata}},
  note         = {Accessed: 2025-04-09}
}

@inproceedings{papadopoulos2007conformal,
  title={Conformal prediction with neural networks},
  author={Papadopoulos, Harris and Vovk, Volodya and Gammerman, Alex},
  booktitle={19th IEEE International Conference on Tools with Artificial Intelligence (ICTAI 2007)},
  volume={2},
  pages={388--395},
  year={2007},
  organization={IEEE}
}

@article{steinberg2024meds_reader,
  title={meds\_reader: A fast and efficient EHR processing library},
  author={Steinberg, Ethan and Wornow, Michael and Bedi, Suhana and Fries, Jason Alan and McDermott, Matthew and Shah, Nigam H},
  journal={arXiv preprint arXiv:2409.09095},
  year={2024}
}

@inproceedings{yang2023pyhealth,
  title={Pyhealth: A deep learning toolkit for healthcare applications},
  author={Yang, Chaoqi and Wu, Zhenbang and Jiang, Patrick and Lin, Zhen and Gao, Junyi and Danek, Benjamin P and Sun, Jimeng},
  booktitle={Proceedings of the 29th ACM SIGKDD Conference on Knowledge Discovery and Data Mining},
  pages={5788--5789},
  year={2023}
}

@article{johnson2019mimic_cxr,
  title={MIMIC-CXR-JPG, a large publicly available database of labeled chest radiographs},
  author={Johnson, Alistair EW and Pollard, Tom J and Greenbaum, Nathaniel R and Lungren, Matthew P and Deng, Chih-ying and Peng, Yifan and Lu, Zhiyong and Mark, Roger G and Berkowitz, Seth J and Horng, Steven},
  journal={arXiv preprint arXiv:1901.07042},
  year={2019}
}

@article{fiotto2024nnsight,
  title={NNsight and NDIF: Democratizing access to open-weight foundation model internals},
  author={Fiotto-Kaufman, Jaden and Loftus, Alexander R and Todd, Eric and Brinkmann, Jannik and Pal, Koyena and Troitskii, Dmitrii and Ripa, Michael and Belfki, Adam and Rager, Can and Juang, Caden and others},
  journal={arXiv preprint arXiv:2407.14561},
  year={2024}
}

@article{rai2024practical_review_survey_mechinterp,
  title={A practical review of mechanistic interpretability for transformer-based language models},
  author={Rai, Daking and Zhou, Yilun and Feng, Shi and Saparov, Abulhair and Yao, Ziyu},
  journal={arXiv preprint arXiv:2407.02646},
  year={2024}
}

@incollection{jain2022huggingface,
  title={Hugging face},
  author={Jain, Shashank Mohan},
  booktitle={Introduction to transformers for NLP: With the hugging face library and models to solve problems},
  pages={51--67},
  year={2022},
  publisher={Springer}
}

@article{wu2024deep_multimodal_missing,
  title={Deep multimodal learning with missing modality: A survey},
  author={Wu, Renjie and Wang, Hu and Chen, Hsiang-Ting and Carneiro, Gustavo},
  journal={arXiv preprint arXiv:2409.07825},
  year={2024}
}

@article{johnson2016mimiciii,
  title={MIMIC-III, a freely accessible critical care database},
  author={Johnson, Alistair EW and Pollard, Tom J and Shen, Lu and Lehman, Li-wei H and Feng, Mengling and Ghassemi, Mohammad and Moody, Benjamin and Szolovits, Peter and Anthony Celi, Leo and Mark, Roger G},
  journal={Scientific data},
  volume={3},
  number={1},
  pages={1--9},
  year={2016},
  publisher={Nature Publishing Group}
}

@article{hripcsak2015omop,
  title={Observational Health Data Sciences and Informatics (OHDSI): opportunities for observational researchers},
  author={Hripcsak, George and Duke, Jon D and Shah, Nigam H and Reich, Christian G and Huser, Vojtech and Schuemie, Martijn J and Suchard, Marc A and Park, Rae Woong and Wong, Ian Chi Kei and Rijnbeek, Peter R and others},
  journal={Studies in health technology and informatics},
  volume={216},
  pages={574},
  year={2015}
}

@article{wang2020mimicextract,
  title={Mimic-extract: A data extraction, preprocessing, and representation pipeline for mimic-iii},
  author={Wang, Shirly and McDermott, Matthew BA and Chauhan, Geeticka and Ghassemi, Marzyeh and Hughes, Michael C and Naumann, Tristan},
  booktitle={Proceedings of the ACM conference on health, inference, and learning},
  pages={222--235},
  year={2020}
}

@article{wornow2023ehrshot,
  title={Ehrshot: An ehr benchmark for few-shot evaluation of foundation models},
  author={Wornow, Michael and Thapa, Rahul and Steinberg, Ethan and Fries, Jason and Shah, Nigam},
  journal={Advances in Neural Information Processing Systems},
  volume={36},
  pages={67125--67137},
  year={2023}
}

@article{knaus1995support,
  title={A controlled trial to improve care for seriously iII hospitalized patients: The study to understand prognoses and preferences for outcomes and risks of treatments (SUPPORT)},
  author={Connors, Alfred F and Dawson, Neal V and Desbiens, Norman A and Fulkerson, William J and Goldman, Lee and Knaus, William A and Lynn, Joanne and Oye, Robert K and Bergner, Marilyn and Damiano, Anne and others},
  journal={Jama},
  volume={274},
  number={20},
  pages={1591--1598},
  year={1995},
  publisher={American Medical Association}
}

@inproceedings{wang2017chestxray14,
  title={Chestx-ray8: Hospital-scale chest x-ray database and benchmarks on weakly-supervised classification and localization of common thorax diseases},
  author={Wang, Xiaosong and Peng, Yifan and Lu, Le and Lu, Zhiyong and Bagheri, Mohammadhadi and Summers, Ronald M},
  booktitle={Proceedings of the IEEE conference on computer vision and pattern recognition},
  pages={2097--2106},
  year={2017}
}

@article{cohen2020covid19cxr, 
  title={Exploring the effect of image enhancement techniques on COVID-19 detection using chest X-ray images},
  author={Rahman, Tawsifur and Khandakar, Amith and Qiblawey, Yazan and Tahir, Anas and Kiranyaz, Serkan and Kashem, Saad Bin Abul and Islam, Mohammad Tariqul and Al Maadeed, Somaya and Zughaier, Susu M and Khan, Muhammad Salman and others},
  journal={Computers in biology and medicine},
  volume={132},
  pages={104319},
  year={2021},
  publisher={Elsevier}
}

@article{kemp2000sleepedf,
  title={Analysis of a sleep-dependent neuronal feedback loop: the slow-wave microcontinuity of the EEG},
  author={Kemp, Bob and Zwinderman, Aeilko H and Tuk, Bert and Kamphuisen, Hilbert AC and Oberye, Josefien JL},
  journal={IEEE Transactions on Biomedical Engineering},
  volume={47},
  number={9},
  pages={1185--1194},
  year={2000},
  publisher={IEEE}
}

@article{khalighi2016isruc,
  title={ISRUC-Sleep: A comprehensive public dataset for sleep researchers},
  author={Khalighi, Sirvan and Sousa, Teresa and Santos, Jos{\'e} Moutinho and Nunes, Urbano},
  journal={Computer methods and programs in biomedicine},
  volume={124},
  pages={180--192},
  year={2016},
  publisher={Elsevier}
}

@article{quan1997shhs,
  title={The sleep heart health study: design, rationale, and methods},
  author={Quan, Stuart F and Howard, Barbara V and Iber, Conrad and Kiley, James P and Nieto, F Javier and O'Connor, George T and Rapoport, David M and Redline, Susan and Robbins, John and Samet, Jonathan M and others},
  journal={Sleep},
  volume={20},
  number={12},
  pages={1077--1085},
  year={1997},
  publisher={Oxford University Press}
}

@article{obeid2016tuab,
  title={The temple university hospital EEG data corpus},
  author={Obeid, Iyad and Picone, Joseph},
  journal={Frontiers in neuroscience},
  volume={10},
  pages={196},
  year={2016},
  publisher={Frontiers Media SA}
}

@article{shah2018tuev,
  title={The temple university hospital seizure detection corpus},
  author={Shah, Vinit and Von Weltin, Eva and Lopez, Silvia and McHugh, James Riley and Veloso, Lillian and Golmohammadi, Meysam and Obeid, Iyad and Picone, Joseph},
  journal={Frontiers in neuroinformatics},
  volume={12},
  pages={83},
  year={2018},
  publisher={Frontiers Media SA}
}

@article{tavara2025prostate_varbench,
  title={Prostate-VarBench: A Benchmark with Interpretable TabNet Framework for Prostate Cancer Variant Classification},
  author={Tavara, Abraham Francisco Arellano and Kumar, Umesh and Pradeepkumar, Jathurshan and Sun, Jimeng},
  journal={arXiv preprint arXiv:2511.09576},
  year={2025}
}

@article{landrum2018clinvar,
  title={ClinVar: improving access to variant interpretations and supporting evidence},
  author={Landrum, Melissa J and Lee, Jennifer M and Benson, Mark and Brown, Garth R and Chao, Chen and Chitipiralla, Shanmuga and Gu, Baoshan and Hart, Jennifer and Hoffman, Douglas and Jang, Wonhee and others},
  journal={Nucleic acids research},
  volume={46},
  number={D1},
  pages={D1062--D1067},
  year={2018},
  publisher={Oxford University Press}
}

@article{tate2019cosmic,
  title={COSMIC: the catalogue of somatic mutations in cancer},
  author={Tate, John G and Bamford, Sally and Jubb, Harry C and Sondka, Zbyslaw and Beare, David M and Bindal, Nidhi and Boutselakis, Harry and Cole, Charlotte G and Creatore, Celestino and Dawson, Elisabeth and others},
  journal={Nucleic acids research},
  volume={47},
  number={D1},
  pages={D941--D947},
  year={2019},
  publisher={Oxford University Press}
}

@article{tcga2015prad,
  title={The cancer genome atlas pan-cancer analysis project},
  author={Weinstein, John N and Collisson, Eric A and Mills, Gordon B and Shaw, Kenna R and Ozenberger, Brad A and Ellrott, Kyle and Shmulevich, Ilya and Sander, Chris and Stuart, Joshua M},
  journal={Nature genetics},
  volume={45},
  number={10},
  pages={1113--1120},
  year={2013},
  publisher={Nature Publishing Group}
}

@article{hochreiter1997lstm,
  title={Long short-term memory},
  author={Hochreiter, Sepp and Schmidhuber, J{\"u}rgen},
  journal={Neural computation},
  volume={9},
  number={8},
  pages={1735--1780},
  year={1997},
  publisher={MIT press}
}

@inproceedings{cho2014gru,
      title={Empirical Evaluation of Gated Recurrent Neural Networks on Sequence Modeling}, 
      author={Junyoung Chung and Caglar Gulcehre and KyungHyun Cho and Yoshua Bengio},
      year={2014},
      eprint={1412.3555},
      archivePrefix={arXiv},
      primaryClass={cs.NE},
      url={https://arxiv.org/abs/1412.3555}, 
}

@article{vaswani2017transformer,
  title={Attention is all you need},
  author={Vaswani, Ashish and Shazeer, Noam and Parmar, Niki and Uszkoreit, Jakob and Jones, Llion and Gomez, Aidan N and Kaiser, {\L}ukasz and Polosukhin, Illia},
  journal={Advances in neural information processing systems},
  volume={30},
  year={2017}
}

@article{lecun1998cnn,
  title={Gradient-based learning applied to document recognition},
  author={LeCun, Yann and Bottou, L{\'e}on and Bengio, Yoshua and Haffner, Patrick},
  journal={Proceedings of the IEEE},
  volume={86},
  number={11},
  pages={2278--2324},
  year={2002},
  publisher={Ieee}
}

@article{bai2018tcn,
  title={An empirical evaluation of generic convolutional and recurrent networks for sequence modeling. arXiv},
  author={Bai, Shaojie and Kolter, J Zico and Koltun, Vladlen},
  journal={arXiv preprint arXiv:1803.01271},
  volume={10},
  year={2018}
}

@article{choi2016retain,
  title={Retain: An interpretable predictive model for healthcare using reverse time attention mechanism},
  author={Choi, Edward and Bahadori, Mohammad Taha and Sun, Jimeng and Kulas, Joshua and Schuetz, Andy and Stewart, Walter},
  journal={Advances in neural information processing systems},
  volume={29},
  year={2016}
}

@inproceedings{gao2020stagenet,
  title={Stagenet: Stage-aware neural networks for health risk prediction},
  author={Gao, Junyi and Xiao, Cao and Wang, Yasha and Tang, Wen and Glass, Lucas M and Sun, Jimeng},
  booktitle={Proceedings of the web conference 2020},
  pages={530--540},
  year={2020}
}

@inproceedings{ma2020adacare,
  title={Adacare: Explainable clinical health status representation learning via scale-adaptive feature extraction and recalibration},
  author={Ma, Liantao and Gao, Junyi and Wang, Yasha and Zhang, Chaohe and Wang, Jiangtao and Ruan, Wenjie and Tang, Wen and Gao, Xin and Ma, Xinyu},
  booktitle={Proceedings of the AAAI Conference on Artificial Intelligence},
  volume={34},
  number={01},
  pages={825--832},
  year={2020}
}

@inproceedings{ma2020concare,
  title={Concare: Personalized clinical feature embedding via capturing the healthcare context},
  author={Ma, Liantao and Zhang, Chaohe and Wang, Yasha and Ruan, Wenjie and Wang, Jiangtao and Tang, Wen and Ma, Xinyu and Gao, Xin and Gao, Junyi},
  booktitle={Proceedings of the AAAI conference on artificial intelligence},
  volume={34},
  number={01},
  pages={833--840},
  year={2020}
}

@inproceedings{zhang2021grasp,
  title={GRASP: generic framework for health status representation learning based on incorporating knowledge from similar patients},
  author={Zhang, Chaohe and Gao, Xin and Ma, Liantao and Wang, Yasha and Wang, Jiangtao and Tang, Wen},
  booktitle={Proceedings of the AAAI conference on artificial intelligence},
  volume={35},
  number={1},
  pages={715--723},
  year={2021}
}

@article{gao2020agent,
  title={Dr. Agent: Clinical predictive model via mimicked second opinions},
  author={Gao, Junyi and Xiao, Cao and Glass, Lucas M and Sun, Jimeng},
  journal={Journal of the American Medical Informatics Association},
  volume={27},
  number={7},
  pages={1084--1091},
  year={2020},
  publisher={Oxford University Press}
}

@article{nguyen2017deepr,
  title={Deepr: a convolutional net for medical records},
  author={Nguyen, Phuoc and Tran, Truyen and Wickramasinghe, Nilmini and Venkatesh, Svetha},
  journal={IEEE journal of biomedical and health informatics},
  volume={21},
  number={1},
  pages={22--30},
  year={2016},
  publisher={IEEE}
}

@inproceedings{jing2023sparcnet,
 title={Development of expert-level classification of seizures and rhythmic and periodic patterns during EEG interpretation},
  author={Jing, Jin and Ge, Wendong and Hong, Shenda and Fernandes, Marta Bento and Lin, Zhen and Yang, Chaoqi and An, Sungtae and Struck, Aaron F and Herlopian, Aline and Karakis, Ioannis and others},
  journal={Neurology},
  volume={100},
  number={17},
  pages={e1750--e1762},
  year={2023},
  publisher={Lippincott Williams \& Wilkins Hagerstown, MD}
}

@article{yang2023contrawr,
  title={Self-supervised EEG representation learning for automatic sleep staging},
  author={Yang, Chaoqi and Xiao, Danica and Westover, M Brandon and Sun, Jimeng},
  journal={arXiv preprint arXiv:2110.15278},
  year={2021}
}

@inproceedings{yang2021safedrug,
  title={Safedrug: Dual molecular graph encoders for recommending effective and safe drug combinations},
  author={Yang, Chaoqi and Xiao, Cao and Ma, Fenglong and Glass, Lucas and Sun, Jimeng},
  journal={arXiv preprint arXiv:2105.02711},
  year={2021}
}

@inproceedings{shang2019gamenet,
  title={Gamenet: Graph augmented memory networks for recommending medication combination},
  author={Shang, Junyuan and Xiao, Cao and Ma, Tengfei and Li, Hongyan and Sun, Jimeng},
  booktitle={proceedings of the AAAI Conference on Artificial Intelligence},
  volume={33},
  number={01},
  pages={1126--1133},
  year={2019}
}

@inproceedings{yang2021micron,
  title={Change matters: Medication change prediction with recurrent residual networks},
  author={Yang, Chaoqi and Xiao, Cao and Glass, Lucas and Sun, Jimeng},
  journal={arXiv preprint arXiv:2105.01876},
  year={2021}
}

@inproceedings{yang2023molerec,
author = {Yang, Nianzu and Zeng, Kaipeng and Wu, Qitian and Yan, Junchi},
title = {MoleRec: Combinatorial Drug Recommendation with Substructure-Aware Molecular Representation Learning},
year = {2023},
isbn = {9781450394161},
publisher = {Association for Computing Machinery},
address = {New York, NY, USA},
url = {https://doi.org/10.1145/3543507.3583872},
doi = {10.1145/3543507.3583872},
booktitle = {Proceedings of the ACM Web Conference 2023},
pages = {4075–4085},
numpages = {11},
location = {Austin, TX, USA},
series = {WWW '23}
}

@article{guevara2024sdoh,
  title={Large language models to identify social determinants of health in electronic health records},
  author={Guevara, Marco and Chen, Shan and Thomas, Spencer and Chaunzwa, Tafadzwa L and Franco, Idalid and Kann, Benjamin H and Moningi, Shalini and Qian, Jack M and Goldstein, Madeleine and Harper, Susan and others},
  journal={NPJ digital medicine},
  volume={7},
  number={1},
  pages={6},
  year={2024},
  publisher={Nature Publishing Group UK London}
}

@article{kingma2014vae,
  title={Auto-encoding variational bayes},
  author={Kingma, Diederik P and Welling, Max},
  journal={arXiv preprint arXiv:1312.6114},
  year={2013}
}

@inproceedings{goodfellow2014gan,
  title={Generative adversarial nets},
  author={Goodfellow, Ian J and Pouget-Abadie, Jean and Mirza, Mehdi and Xu, Bing and Warde-Farley, David and Ozair, Sherjil and Courville, Aaron and Bengio, Yoshua},
  journal={Advances in neural information processing systems},
  volume={27},
  year={2014}
}

@inproceedings{velickovic2018gat,
  title={Graph attention networks},
  author={Veli{\v{c}}kovi{\'c}, Petar and Cucurull, Guillem and Casanova, Arantxa and Romero, Adriana and Lio, Pietro and Bengio, Yoshua},
  journal={arXiv preprint arXiv:1710.10903},
  year={2017}
}

@inproceedings{kipf2017gcn,
title={Semi-supervised classification with graph convolutional networks},
  author={Kipf, TN},
  journal={arXiv preprint arXiv:1609.02907},
  year={2016}
}

@inproceedings{devlin2019bert,
   title={Bert: Pre-training of deep bidirectional transformers for language understanding},
  author={Devlin, Jacob and Chang, Ming-Wei and Lee, Kenton and Toutanova, Kristina},
  booktitle={Proceedings of the 2019 conference of the North American chapter of the association for computational linguistics: human language technologies, volume 1 (long and short papers)},
  pages={4171--4186},
  year={2019}
}

@inproceedings{alsentzer2019clinicalbert,
  title={Publicly available clinical BERT embeddings},
  author={Alsentzer, Emily and Murphy, John and Boag, William and Weng, Wei-Hung and Jindi, Di and Naumann, Tristan and McDermott, Matthew},
  booktitle={Proceedings of the 2nd clinical natural language processing workshop},
  pages={72--78},
  year={2019}
}

@misc{landes2025integrationlargelanguagemodels_sdoh,
      title={Integration of Large Language Models and Traditional Deep Learning for Social Determinants of Health Prediction}, 
      author={Paul Landes and Jimeng Sun and Adam Cross},
      year={2025},
      eprint={2505.04655},
      archivePrefix={arXiv},
      primaryClass={cs.CL},
      url={https://arxiv.org/abs/2505.04655}, 
}

@inproceedings{hassan2024reproducibility_dependency,
  title={Reproducibility debt: Challenges and future pathways},
  author={Hassan, Zara and Treude, Christoph and Norrish, Michael and Williams, Graham and Potanin, Alex},
  booktitle={Companion Proceedings of the 32nd ACM International Conference on the Foundations of Software Engineering},
  pages={462--466},
  year={2024}
}

@misc{semmelrock2023reproducibilitymachinelearningdrivenresearch,
      title={Reproducibility in Machine Learning-Driven Research}, 
      author={Harald Semmelrock and Simone Kopeinik and Dieter Theiler and Tony Ross-Hellauer and Dominik Kowald},
      year={2023},
      eprint={2307.10320},
      archivePrefix={arXiv},
      primaryClass={cs.LG},
      url={https://arxiv.org/abs/2307.10320}, 
}

@inproceedings{zadrozny2001calibration,
  title={Obtaining calibrated probability estimates from decision trees and naive bayesian classifiers},
  author={Zadrozny, Bianca and Elkan, Charles},
  booktitle={Icml},
  volume={1},
  number={05},
  year={2001}
}

@inproceedings{kull2019dirichlet,
  title={Beyond temperature scaling: Obtaining well-calibrated multi-class probabilities with dirichlet calibration},
  author={Kull, Meelis and Perello Nieto, Miquel and K{\"a}ngsepp, Markus and Silva Filho, Telmo and Song, Hao and Flach, Peter},
  journal={Advances in neural information processing systems},
  volume={32},
  year={2019}
}

@article{lin2023kcal,
  title={Taking a step back with kcal: Multi-class kernel-based calibration for deep neural networks},
  author={Lin, Zhen and Trivedi, Shubhendu and Sun, Jimeng},
  journal={arXiv preprint arXiv:2202.07679},
  year={2022}
}

@article{sadinle2019label,
  title={Least ambiguous set-valued classifiers with bounded error levels},
  author={Sadinle, Mauricio and Lei, Jing and Wasserman, Larry},
  journal={Journal of the American Statistical Association},
  volume={114},
  number={525},
  pages={223--234},
  year={2019},
  publisher={Taylor \& Francis}
}

@inproceedings{lin2022scrib,
  title={Scrib: set-classifier with class-specific risk bounds for blackbox models},
  author={Lin, Zhen and Glass, Lucas and Westover, M Brandon and Xiao, Cao and Sun, Jimeng},
  booktitle={Proceedings of the AAAI Conference on Artificial Intelligence},
  volume={36},
  number={7},
  pages={7497--7505},
  year={2022}
}

@inproceedings{lin2023favmac,
  title={Fast online value-maximizing prediction sets with conformal cost control},
  author={Lin, Zhen and Trivedi, Shubhendu and Xiao, Cao and Sun, Jimeng},
  booktitle={International Conference on Machine Learning},
  pages={21182--21203},
  year={2023},
  organization={PMLR}
}

@article{tibshirani2019conformal,
  title={Conformal prediction under covariate shift},
  author={Tibshirani, Ryan J and Foygel Barber, Rina and Candes, Emmanuel and Ramdas, Aaditya},
  journal={Advances in neural information processing systems},
  volume={32},
  year={2019}
}

@article{laghuvarapu2023codrug,
  title={Codrug: Conformal drug property prediction with density estimation under covariate shift},
  author={Laghuvarapu, Siddhartha and Lin, Zhen and Sun, Jimeng},
  journal={Advances in Neural Information Processing Systems},
  volume={36},
  pages={37728--37747},
  year={2023}
}

@inproceedings{simonyan2014saliency,
  title={Deep inside convolutional networks: Visualising image classification models and saliency maps},
  author={Simonyan, Karen and Vedaldi, Andrea and Zisserman, Andrew},
  journal={arXiv preprint arXiv:1312.6034},
  year={2013}
}

@inproceedings{sundararajan2017integrated,
  title={Axiomatic attribution for deep networks},
  author={Sundararajan, Mukund and Taly, Ankur and Yan, Qiqi},
  booktitle={International conference on machine learning},
  pages={3319--3328},
  year={2017},
  organization={PMLR}
}

@inproceedings{shrikumar2017deeplift,
  title={Learning important features through propagating activation differences},
  author={Shrikumar, Avanti and Greenside, Peyton and Kundaje, Anshul},
  booktitle={International conference on machine learning},
  pages={3145--3153},
  year={2017},
  organization={PMlR}
}

@inproceedings{ribeiro2016lime,
  title={" Why should i trust you?" Explaining the predictions of any classifier},
  author={Ribeiro, Marco Tulio and Singh, Sameer and Guestrin, Carlos},
  booktitle={Proceedings of the 22nd ACM SIGKDD international conference on knowledge discovery and data mining},
  pages={1135--1144},
  year={2016}
}

@article{rumelhart1986learning,
  title={Learning representations by back-propagating errors},
  author={Rumelhart, David E and Hinton, Geoffrey E and Williams, Ronald J},
  journal={nature},
  volume={323},
  number={6088},
  pages={533--536},
  year={1986},
  publisher={Nature Publishing Group UK London}
}

@article{PhysioNet-challenge-2020-1.0.2_Cardiology,
  author = {{Perez Alday}, Erick Andres and Gu, Annie and Shah, Amit and Liu, Chengyu and Sharma, Ashish and Seyedi, Salman and {Bahrami Rad}, Ali and Reyna, Matthew and Clifford, Gari},
  title = {{Classification of 12-lead ECGs: The PhysioNet/Computing in Cardiology Challenge 2020}},
  journal = {{PhysioNet}},
  year = {2022},
  month = jul,
  note = {Version 1.0.2},
  doi = {10.13026/dvyd-kd57},
  url = {https://doi.org/10.13026/dvyd-kd57}
}

@misc{ali2024buetmultidiseaseheartsound_bmd_hs,
      title={BUET Multi-disease Heart Sound Dataset: A Comprehensive Auscultation Dataset for Developing Computer-Aided Diagnostic Systems}, 
      author={Shams Nafisa Ali and Afia Zahin and Samiul Based Shuvo and Nusrat Binta Nizam and Shoyad Ibn Sabur Khan Nuhash and Sayeed Sajjad Razin and S. M. Sakeef Sani and Farihin Rahman and Nawshad Binta Nizam and Farhat Binte Azam and Rakib Hossen and Sumaiya Ohab and Nawsabah Noor and Taufiq Hasan},
      year={2024},
      eprint={2409.00724},
      archivePrefix={arXiv},
      primaryClass={eess.SP},
      url={https://arxiv.org/abs/2409.00724}, 
}

@article{PhysioNet-dreamt-2.1.0,
  author = {Wang, Ke and Yang, Jiamu and Shetty, Ayush and Dunn, Jessilyn},
  title = {{DREAMT: Dataset for Real-time sleep stage EstimAtion using Multisensor wearable Technology}},
  journal = {{PhysioNet}},
  year = {2025},
  month = apr,
  note = {Version 2.1.0},
  doi = {10.13026/7r9r-7r24},
  url = {https://doi.org/10.13026/7r9r-7r24}
}

@inproceedings{landes2023deepzensols,
  title={DeepZensols: A deep learning natural language processing framework for experimentation and reproducibility},
  author={Landes, Paul and Di Eugenio, Barbara and Caragea, Cornelia},
  booktitle={Proceedings of the 3rd Workshop for Natural Language Processing Open Source Software (NLP-OSS 2023)},
  pages={141--146},
  year={2023}
}

@article{cina2025we_need_XAI,
  title={Why we do need explainable ai for healthcare},
  author={Cin{\`a}, Giovanni and R{\"o}ber, Tabea E and Goedhart, Rob and Birbil, {\c{S}} {\.I}lker},
  journal={Diagnostic and Prognostic Research},
  volume={9},
  number={1},
  pages={24},
  year={2025},
  publisher={Springer}
}

@article{he2025survey,
  title={A survey on uncertainty quantification methods for deep learning},
  author={He, Wenchong and Jiang, Zhe and Xiao, Tingsong and Xu, Zelin and Li, Yukun},
  journal={ACM Computing Surveys},
  year={2025},
  publisher={ACM New York, NY}
}
\bibliographystyle{icml2026}

\newpage
\appendix
\onecolumn
\section{Implementation Details}
All code is shared in the supplementary material, specifically with our Pandas and PyHealth 1.16 comparisons. We share our code through \url{https://anonymous.4open.science/r/PyHealth-CB8D/}, and recommend installing PyHealth 1.16 through \texttt{pip install pyhealth==1.16}. To reimplement our MEDS baseline, please install \texttt{meds\_etl} and \texttt{meds\_reader} through \texttt{pip install meds\_etl meds\_reader} respectively. For MEDS\_etl, we use the default settings as denoted by their GitHub settings, which assumes 100 shards for caching. We vary the number of processors in conjunction with our framework to better understand how each framework scales across different compute levels. 

\section{Explicit Benchmark Numbers for ML Task Processsing} \label{appdx: pyhealth table perf}
We present exact performance numbers here for our three performance benchmarks from loading all of the raw data to final task processing in Tables \ref{tab:drug-benchmark-perf}, \ref{tab:los-benchmark-perf}, and \ref{tab:mortality-benchmark-perf}. We observe that PyHealth 2.0 exhibits the lowest peak memory usage while generally being comparable or even faster than PyHealth 1.16's approach despite PyHealth 1.16 loading everything in memory for parallel processing.
\begin{table}[h]
\centering
\caption{Performance comparison for in-hospital mortality prediction on MIMIC-IV.
Wall time in seconds, memory in GB. Best results per metric in bold.
$\dagger$\,PyHealth~2.0 laptop rows benchmarked at 4 and 8 workers only.}
\label{tab:mortality-benchmark-perf}
\small
\begin{tabular}{l|ccccc}
\toprule
\multirow{2}{*}{\textbf{Method}} & \multicolumn{5}{c}{\textbf{Number of Workers}} \\
& 1 & 4 & 8 & 12 & 16 \\
\midrule
\multicolumn{6}{l}{\textit{Wall Time (seconds)}} \\
Pandas                                  & 93,708          & \textcolor{red}{\ding{55}} & \textcolor{red}{\ding{55}} & \textcolor{red}{\ding{55}} & \textcolor{red}{\ding{55}} \\
PyHealth 1.16                           & \textbf{6,841}  & 3,174          & 2,543          & \textbf{2,295} & \textbf{2,235} \\
MEDS                                    & 17,202          & 10,672         & 10,117         & 10,707         & 9,694          \\
PyHealth 2.0 (Large WS)                 & 17,671          & 4,605          & 2,975          & 2,551          & 2,385          \\
PyHealth 2.0 (MacBook Pro M2 Pro)$\dagger$ & ---          & \textbf{2,010} & \textbf{1,357} & ---            & ---            \\
PyHealth 2.0 (ASUS Zephyrus G16)$\dagger$ & ---          & 2,617          & 1,605          & ---            & ---            \\
\midrule
\multicolumn{6}{l}{\textit{Peak Memory (GB)}} \\
Pandas                                  & 49.23           & \textcolor{red}{\ding{55}} & \textcolor{red}{\ding{55}} & \textcolor{red}{\ding{55}} & \textcolor{red}{\ding{55}} \\
PyHealth 1.16                           & 146.23          & 187.21         & 275.73         & 363.13         & 457.42         \\
MEDS                                    & 58.07           & 60.00          & 61.48          & 61.62          & 65.19          \\
PyHealth 2.0 (Large WS)                 & \textbf{17.48}  & 23.39          & 23.57          & \textbf{24.19} & \textbf{28.70} \\
PyHealth 2.0 (MacBook Pro M2 Pro)$\dagger$ & ---          & \textbf{5.31}  & \textbf{7.43}  & ---            & ---            \\
PyHealth 2.0 (ASUS Zephyrus G16)$\dagger$ & ---          & 19.15          & 19.67          & ---            & ---            \\
\bottomrule
\end{tabular}
\end{table}

\begin{table}[h]
\centering
\caption{Performance comparison for drug recommendation on MIMIC-IV (excluding lab events). Wall time in seconds, memory in GB. Best results per metric in bold.}
\label{tab:drug-benchmark-perf}
\small
\begin{tabular}{l|ccccc}
\toprule
\multirow{2}{*}{\textbf{Method}} & \multicolumn{5}{c}{\textbf{Number of Workers}} \\
& 1 & 4 & 8 & 12 & 16 \\
\midrule
\multicolumn{6}{l}{\textit{Wall Time (seconds)}} \\
Pandas & 4,725 & \textcolor{red}{\ding{55}} & \textcolor{red}{\ding{55}} & \textcolor{red}{\ding{55}} & \textcolor{red}{\ding{55}} \\
PyHealth 1.16 & \textbf{1,841} & 981 & 857 & 860 & 802 \\
MEDS & 17,544 & 10,477 & 10,398 & 10,968 & 9,840 \\
PyHealth 2.0 & 2,375 & \textbf{841} & \textbf{610} & \textbf{581} & \textbf{569} \\
\midrule
\multicolumn{6}{l}{\textit{Peak Memory (GB)}} \\
Pandas & 10.41 & \textcolor{red}{\ding{55}} & \textcolor{red}{\ding{55}} & \textcolor{red}{\ding{55}} & \textcolor{red}{\ding{55}} \\
PyHealth 1.16 & 56.62 & 83.24 & 125.22 & 167.84 & 211.62 \\
MEDS & 54.45 & 61.66 & 66.47 & 65.26 & 70.72 \\
PyHealth 2.0 & \textbf{6.63} & \textbf{8.86} & \textbf{11.80} & \textbf{15.04} & \textbf{18.50} \\
\bottomrule
\end{tabular}
\end{table}

\begin{table}[h!]
\centering
\caption{Performance comparison for length of stay prediction on MIMIC-IV (excluding lab events). Wall time in seconds, memory in GB. Best results per metric in bold.}
\label{tab:los-benchmark-perf}
\small
\begin{tabular}{l|ccccc}
\toprule
\multirow{2}{*}{\textbf{Method}} & \multicolumn{5}{c}{\textbf{Number of Workers}} \\
& 1 & 4 & 8 & 12 & 16 \\
\midrule
\multicolumn{6}{l}{\textit{Wall Time (seconds)}} \\
Pandas & 10,269 & \textcolor{red}{\ding{55}} & \textcolor{red}{\ding{55}} & \textcolor{red}{\ding{55}} & \textcolor{red}{\ding{55}} \\
PyHealth 1.16 & \textbf{4,434} & \textbf{2,316} & 2,054 & 1,965 & 1,922 \\
MEDS & 21,236 & 11,947 & 9,003 & 8,077 & 8,135 \\
PyHealth 2.0 & 14,228 & 2,484 & \textbf{1,396} & \textbf{1,176} & \textbf{1,096} \\
\midrule
\multicolumn{6}{l}{\textit{Peak Memory (GB)}} \\
Pandas & 11.40 & \textcolor{red}{\ding{55}} & \textcolor{red}{\ding{55}} & \textcolor{red}{\ding{55}} & \textcolor{red}{\ding{55}} \\
PyHealth 1.16 & 56.60 & 80.89 & 123.76 & 165.79 & 209.93 \\
MEDS & 57.95 & 61.82 & 64.66 & 69.30 & 71.71 \\
PyHealth 2.0 & \textbf{6.40} & \textbf{8.46} & \textbf{11.31} & \textbf{14.49} & \textbf{17.83} \\
\bottomrule
\end{tabular}
\end{table}

\section{Dataloading and Caching Implementation Details} \label{appdx : dataloading and caching details}
The dataloading pipeline consists of three cached stages: table-joining, task-transformation, processor-transformation. 

Table-joining merges multiple CSVs into a single comprehensive event table. We use Dask \cite{rocklin2015dask} to join all tables due to its out-of-core joining and sorting capabilities, wrapped by Narwhals to ensure a consistent, Polars-like API for PyHealth. The output is sorted by \texttt{patient\_id} and cached as Parquet files, allowing downstream processes to utilize Parquet min-max statistics for faster grouping.

Task-transformation groups events by \texttt{patient\_id} to create samples based on specific PyHealth tasks. We use Polars to process batches of 128 consecutive patients; this approach optimizes speed by leveraging data locality and row-group statistics while keeping the memory usage low. To prevent thread starvation in multi-worker settings, the Polars thread pool count is calibrated per worker. Resulting samples are cached in LitData binary format \cite{litdata2023}.

Processor-transformation converts the generated samples into tensors. The output is cached in LitData binary format. During PyTorch training, we employ LitData's \texttt{StreamingDataset} \cite{litdata2023} to enable efficient data loading, subsampling, shuffling, and splitting.

\section{Dataloading and Caching Performance Comparison} \label{appdx : dataloading and caching perf}
In principle, most datasets are cached after processing, eliminating the high computational costs of initial data loading from raw .csv files. Once processed, iteratively exploring patient data and constructing new ML tasks becomes relatively fast. Table \ref{tab:cache_times_patient_data_access} examines the time required to join all MIMIC-IV \citep{johnson2023mimic4} tables and format them into patient representations with all events. Note that this excludes machine learning task feature engineering and filtering of patients without specific events.

Remarkably, our naive Pandas solution proves remarkably fastest for table joins and caching, with relatively fast patient access times. Nonetheless, PyHealth 2.0 serves as the nearest competitive alternative. A key design change that leads to slower random patient access compared to its predecessor PyHealth 1.16 and related software MEDS\_reader \citep{steinberg2024meds_reader} is the elimination of patient index assumptions, removing fast-access maps of patient data. This design change accommodates imaging and signal datasets where patient IDs are not necessarily available. Compared to Pandas, PyHealth 2.0 stores much of the patient data on disk rather than directly in memory, resulting in slightly slower patient access times. Nevertheless, patient access times remain in the millisecond range, enabling fundamentally fast patient exploration. Future work could explore efficient patient data mapping strategies. This benchmark uses the same code structure shown in Table \ref{tab:loc_comparison}.

\begin{table}[h]
\centering
\caption{\textbf{Dataloading, cache, and random patient access times.} *We note that part of our dataloading time with MEDS\_reader \citep{steinberg2024meds_reader} includes the conversion time from PyHealth 1.16, making it ultimately slower in this comparison to PyHealth 1.16. Nonetheless, patient access times are remarkably fast here in all cases, being measured in milliseconds. Please note that this patient random access time does not necessarily result in worse as all patients are loaded serially in training. }
\label{tab:cache_times_patient_data_access}
\resizebox{1.0\textwidth}{!}{%
\begin{tabular}{@{}lccccc@{}}
\toprule
& Pandas & PyHealth 1.16 & MEDS ETL + MEDS\_Reader & PyHealth 1.16 + MEDS\_Reader & PyHealth 2.0 \\
\midrule
Load \& Cache Time (s) & 740.97 & 3548.90 & 10117.04 & 3766.31* & 1093.38 \\
Patient Access Time (ms) & 168.23 & 0.01 & 0.01 & 0.02 & 208.85 \\
\bottomrule
\end{tabular}
}
\end{table}

\section{Example Use Cases} \label{appdx: interp and multimodality}
To demonstrate PyHealth 2.0's extensive feature set, we present qualitative examples showcasing major framework capabilities: (1) multimodal patient dataloading and (2) integrated interpretability and uncertainty quantification for model evaluation.

\begin{figure*}[h!]
    \centering
    \includegraphics[width=0.95\textwidth]{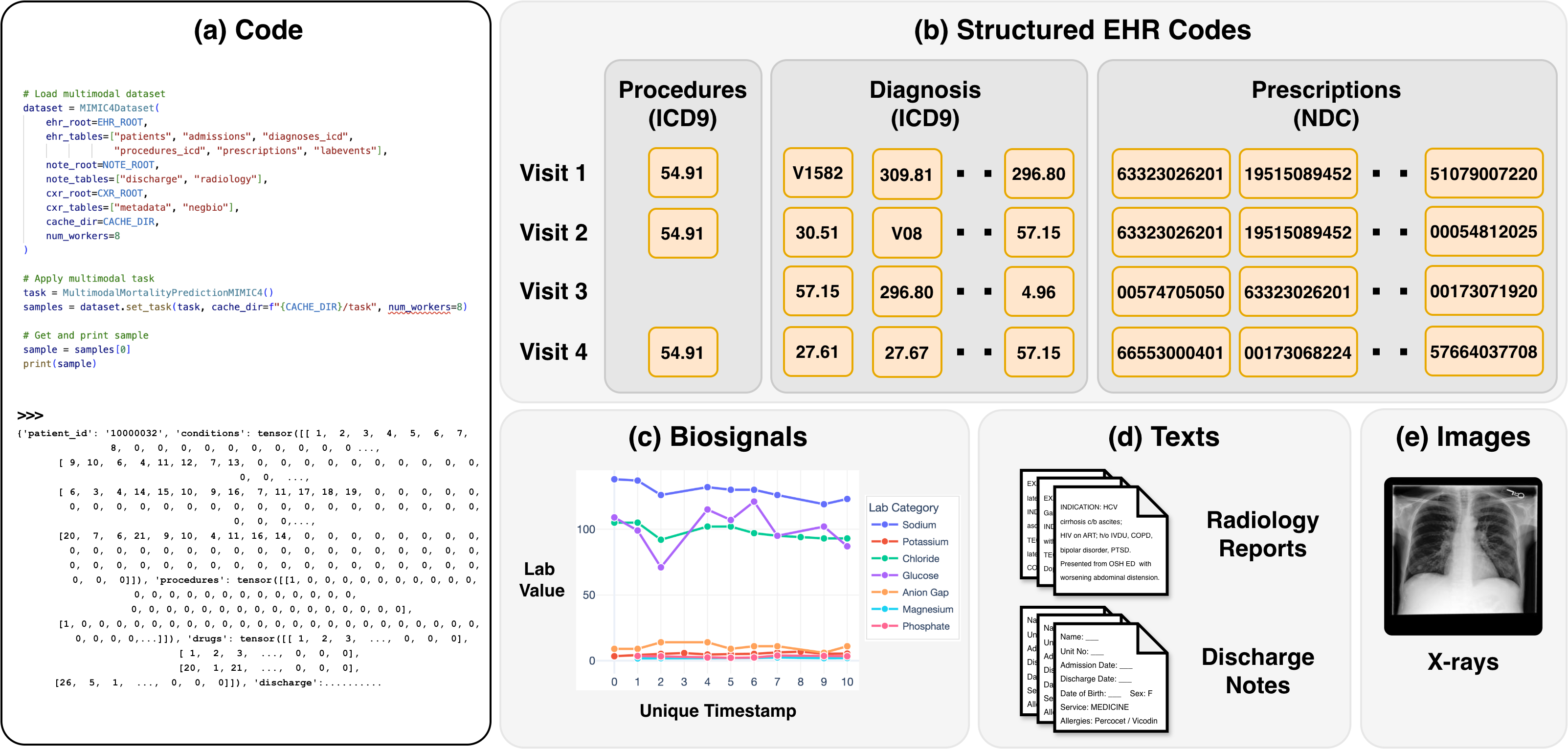}
    \caption{\textbf{PyHealth 2.0 directly supports loading multimodal data.} With effectively only 5 lines of code, PyHealth 2.0 now supports the ability to work with (b) structured EHR codes, (c) biosignals, (d) clinical notes, (e) and X-rays on MIMIC-IV data \citep{johnson2023mimic4}. }
    \label{fig:multimodal data loading}
   \vspace{-0.3cm}
\end{figure*}
\begin{figure*}[h!]
    \centering
    \includegraphics[width=0.95\textwidth]{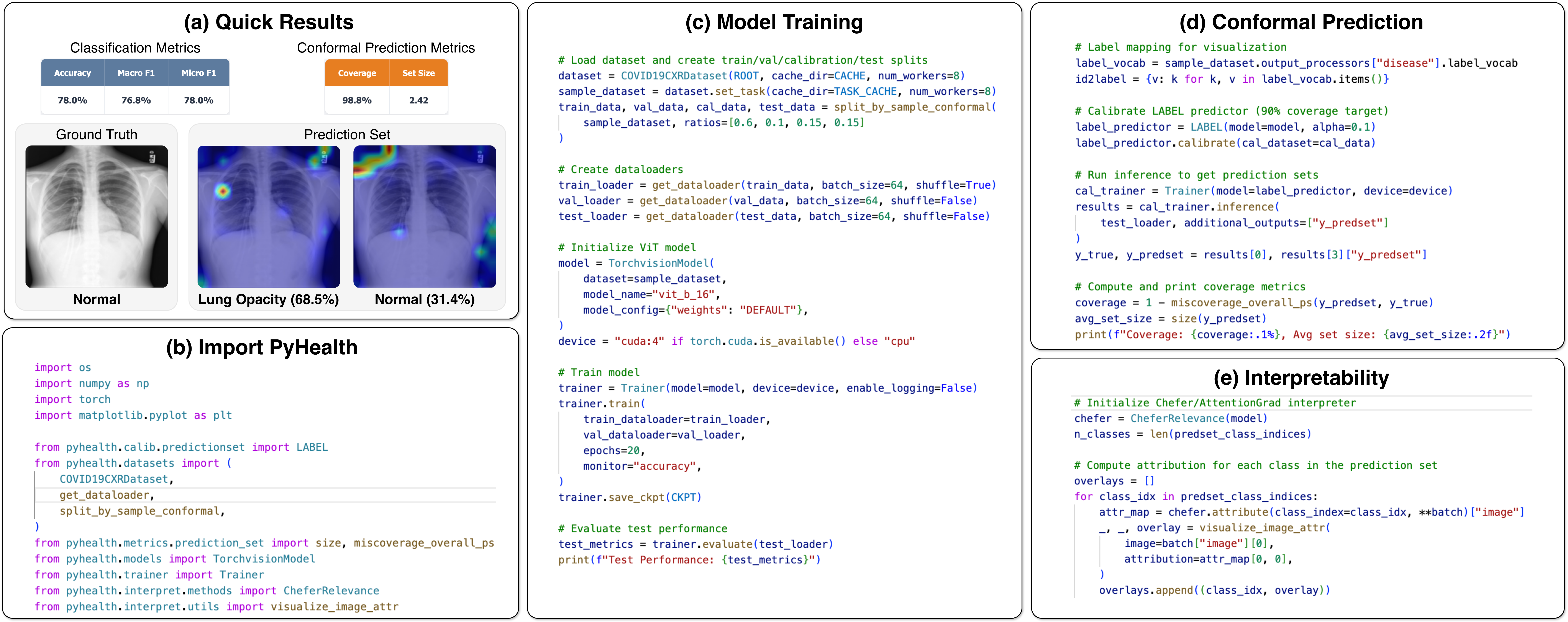}
    \caption{\textbf{PyHealth 2.0 supports various interpretability and uncertainty quantification deployment techniques.} To produce our results (a), we import PyHealth (b), train our model (c), finish conformal prediction (d), and run an interpretability visualization (e).}
    \vspace{-0.5cm}
    \label{fig:cxr_example}
\end{figure*}

\textbf{Multimodal dataloading.} In Figure \ref{fig:multimodal data loading}, we show how we can construct a dataset and explore an entire patient's extensive profile in MIMIC-IV. The MIMIC4Dataset.set\_task() function is called on the MultiModalMortalityPrediction task object here, generating a sequence of multimodal events that quickly enables us to generate the visualizations in the rest of Figure \ref{fig:multimodal data loading}.

\textbf{Interpretability and uncertainty quantification.} We fine-tune a Vision Transformer (ViT) on the COVID-19 Chest X-ray dataset \citep{cohen2020covid19cxr} to illustrate PyHealth's capacity for deeper model evaluation beyond standard predictive metrics. Clinical X-ray classification demands interpretable predictions to support diagnostic decision-making, while uncertainty quantification provides theoretical guarantees essential for building deployment trust. We configure conformal prediction with $\alpha=0.01$ and evaluate empirical coverage through \texttt{pyhealth.metrics}. Subsequently, we apply \texttt{pyhealth.interpretability} to analyze the fine-tuned ViT's decision-making robustness.

\textbf{Conformal interpretability results.} Our vanilla conformal prediction \citep{angelopoulos2023conformal, papadopoulos2007conformal} achieves approximately 99\% empirical coverage on the test set, consistent with the configured $\alpha=0.01$ miscoverage rate. However, interpretability analysis using AttentionGrad \citep{chefer2021genericattentionmodelexplainabilityinterpreting_attngrad} reveals spurious feature dependence: the ViT incorrectly predicts "lung opacity" for a normal patient X-ray by primarily attending to annotation text in the image corner rather than anatomical features. While the conformal predictor correctly captures the ground truth as the second-most-likely class, the attention maps confirm reliance on spurious correlations here.
%
%

\section{Benchmark Results}
\label{app:benchmarks}

To complement the data-processing benchmarks in the main paper, we provide
downstream modeling results across every modality supported by PyHealth~2.0
(EHR, imaging, text, and biosignals), together with interpretability and
uncertainty-quantification benchmarks. These results are intended as evidence
that PyHealth~2.0's standardized task schemas and default configurations yield
reasonable, reproducible baselines out of the box.

Unless otherwise noted, every model in this appendix is trained for
\textbf{20 epochs} with \textbf{AdamW} (learning rate $1\times10^{-4}$) using
the framework's default hyperparameters. We deliberately perform
\emph{no} hyperparameter tuning and \emph{no} pretraining: the goal is to
characterize default reproducible behavior rather than state-of-the-art
performance.

\subsection{Datasets and Tasks}
\label{app:datasets}

Benchmarks were collected on two compute servers. Table~\ref{tab:dataset-overview}
summarizes the datasets and clinical tasks evaluated on each.

\begin{table}[!ht]
  \centering
  \caption{Datasets and tasks benchmarked, grouped by compute server.
  Datasets: MIMIC-III~\cite{johnson2016mimiciii},
  MIMIC-IV~\cite{johnson2023mimic4}, eICU~\cite{pollard2018eicu},
  Covid19CXR~\cite{cohen2020covid19cxr}, TUAB~\cite{obeid2016tuab}, and
  TUEV~\cite{shah2018tuev}.}
  \label{tab:dataset-overview}
  \begin{tabular}{ll}
    \toprule
    \textbf{Dataset} & \textbf{Task} \\
    \midrule
    \multicolumn{2}{l}{\emph{Server 1}} \\
    \midrule
    MIMIC-III   & Mortality Prediction \\
    MIMIC-III   & Readmission Prediction \\
    Covid19CXR  & Classification \\
    MIMIC-IV    & Mortality Prediction \\
    MIMIC-IV    & Length of Stay Prediction \\
    MIMIC-IV    & Drug Recommendation \\
    MIMIC-IV    & Readmission Prediction \\
    \midrule
    \multicolumn{2}{l}{\emph{Server 3}} \\
    \midrule
    eICU                    & Drug Recommendation \\
    eICU                    & Length of Stay \\
    eICU                    & Mortality Prediction \\
    eICU                    & Readmission Prediction \\
    Medical Transcriptions  & Classification \\
    TUH EEG (TUAB)          & Abnormal EEG Detection \\
    TUH EEG (TUEV)          & EEG Event Classification \\
    \bottomrule
  \end{tabular}
\end{table}

\clearpage
\subsection{EHR Modeling}
\label{app:ehr}

\begin{table}[!ht]
  \centering
  \caption{Mortality prediction on MIMIC-III~\cite{johnson2016mimiciii}.}
  \label{tab:mimic3-mortality}
  \begin{tabular}{lccccc}
    \toprule
    \textbf{Model} & \textbf{AUROC} & \textbf{PRAUC} & \textbf{F1} & \textbf{Accuracy} & \textbf{Loss} \\
    \midrule
    RNN          & 0.5778 & 0.1378 & 0.0000 & 0.8967 & 0.3338 \\
    StageNet     & 0.5635 & 0.1323 & 0.0000 & 0.8967 & 0.3336 \\
    Adacare      & 0.5505 & 0.1520 & 0.0000 & 0.8967 & 0.3317 \\
    Retain       & 0.5493 & 0.1175 & 0.0000 & 0.8967 & 0.3419 \\
    Transformer  & 0.5415 & 0.1268 & 0.1152 & 0.8236 & 1.5007 \\
    ConCare      & 0.5043 & 0.1080 & 0.0000 & 0.8967 & 0.3381 \\
    \bottomrule
  \end{tabular}
\end{table}

\begin{table}[!ht]
  \centering
  \caption{Readmission prediction on MIMIC-III~\cite{johnson2016mimiciii}.}
  \label{tab:mimic3-readmission}
  \begin{tabular}{lccccc}
    \toprule
    \textbf{Model} & \textbf{AUROC} & \textbf{PRAUC} & \textbf{F1} & \textbf{Accuracy} & \textbf{Loss} \\
    \midrule
    ConCare      & 0.5576 & 0.2100 & 0.0000 & 0.8278 & 0.4630 \\
    RNN          & 0.5549 & 0.1955 & 0.0000 & 0.8257 & 0.4633 \\
    Adacare      & 0.5500 & 0.2139 & 0.0000 & 0.8278 & 0.4584 \\
    Retain       & 0.5500 & 0.1967 & 0.0000 & 0.8236 & 0.4707 \\
    TCN          & 0.5423 & 0.2075 & 0.1370 & 0.8027 & 0.5157 \\
    Transformer  & 0.5296 & 0.1785 & 0.0221 & 0.8152 & 0.5173 \\
    \bottomrule
  \end{tabular}
\end{table}

\begin{table}[!ht]
  \centering
  \caption{Mortality prediction on MIMIC-IV~\cite{johnson2023mimic4}.}
  \label{tab:mimic4-mortality}
  \begin{tabular}{lcc}
    \toprule
    \textbf{Model} & \textbf{AUROC} & \textbf{PRAUC} \\
    \midrule
    Deepr        & 0.7547 & 0.0692 \\
    RNN          & 0.7481 & 0.0673 \\
    StageNet     & 0.7422 & 0.0791 \\
    Retain       & 0.7250 & 0.0588 \\
    Transformer  & 0.7109 & 0.0502 \\
    ConCare      & 0.6879 & 0.0540 \\
    Adacare      & 0.6653 & 0.0425 \\
    EHRMamba     & 0.6224 & 0.0456 \\
    \bottomrule
  \end{tabular}
\end{table}

\begin{table}[!ht]
  \centering
  \caption{Length of stay prediction on MIMIC-IV~\cite{johnson2023mimic4}.}
  \label{tab:mimic4-los}
  \begin{tabular}{lccc}
    \toprule
    \textbf{Model} & \textbf{Accuracy} & \textbf{F1 Macro} & \textbf{Loss} \\
    \midrule
    RNN          & 0.4567 & 0.4050 & 1.3752 \\
    Retain       & 0.4341 & 0.3792 & 1.4364 \\
    Transformer  & 0.4202 & 0.3529 & 1.5486 \\
    Deepr        & 0.4144 & 0.3511 & 1.4949 \\
    EHRMamba     & 0.3995 & 0.3349 & 1.5475 \\
    Adacare      & 0.3825 & 0.3142 & 1.6043 \\
    \bottomrule
  \end{tabular}
\end{table}

\begin{table}[!ht]
  \centering
  \caption{Drug recommendation on MIMIC-IV~\cite{johnson2023mimic4}.}
  \label{tab:mimic4-drugrec}
  \begin{tabular}{lccc}
    \toprule
    \textbf{Model} & \textbf{Jaccard} & \textbf{F1 Samples} & \textbf{Loss} \\
    \midrule
    EHRMamba     & 0.4915 & 0.6503 & 0.0602 \\
    Deepr        & 0.4813 & 0.6408 & 0.0616 \\
    Transformer  & 0.4661 & 0.6266 & 0.0632 \\
    Adacare      & 0.4659 & 0.6268 & 0.0629 \\
    RNN          & 0.4301 & 0.5929 & 0.0677 \\
    GameNet      & 0.4278 & 0.5910 & 0.0690 \\
    Retain       & 0.4234 & 0.5873 & 0.0701 \\
    \bottomrule
  \end{tabular}
\end{table}

\begin{table}[!ht]
  \centering
  \caption{Mortality prediction on eICU~\cite{pollard2018eicu}.}
  \label{tab:eicu-mortality}
  \begin{tabular}{lccccc}
    \toprule
    \textbf{Model} & \textbf{AUROC} & \textbf{PRAUC} & \textbf{F1} & \textbf{Accuracy} & \textbf{Loss} \\
    \midrule
    Retain       & 0.6836 & 0.1807 & 0.0071 & 0.9183 & 0.2656 \\
    Transformer  & 0.6696 & 0.1847 & 0.0530 & 0.9159 & 0.2788 \\
    RNN          & 0.6693 & 0.2136 & 0.1859 & 0.9150 & 0.2850 \\
    ConCare      & 0.6306 & 0.1448 & 0.0000 & 0.9180 & 0.2773 \\
    Adacare      & 0.6248 & 0.1315 & 0.0000 & 0.9180 & 0.2781 \\
    \bottomrule
  \end{tabular}
\end{table}

\begin{table}[!ht]
  \centering
  \caption{Length of stay prediction on eICU~\cite{pollard2018eicu}.}
  \label{tab:eicu-los}
  \begin{tabular}{lccc}
    \toprule
    \textbf{Model} & \textbf{Accuracy} & \textbf{F1 Macro} & \textbf{Loss} \\
    \midrule
    RNN          & 0.3912 & 0.2420 & 1.5503 \\
    Retain       & 0.3796 & 0.1955 & 1.5876 \\
    Transformer  & 0.3696 & 0.1719 & 1.6659 \\
    Adacare      & 0.3547 & 0.1609 & 1.6569 \\
    ConCare      & 0.3491 & 0.1545 & 1.6655 \\
    \bottomrule
  \end{tabular}
\end{table}

\begin{table}[!ht]
  \centering
  \caption{Readmission prediction on eICU~\cite{pollard2018eicu}.}
  \label{tab:eicu-readmission}
  \begin{tabular}{lccccc}
    \toprule
    \textbf{Model} & \textbf{AUROC} & \textbf{PRAUC} & \textbf{F1} & \textbf{Accuracy} & \textbf{Loss} \\
    \midrule
    Transformer  & 0.8151 & 0.8570 & 0.7497 & 0.7353 & 0.5357 \\
    RNN          & 0.8071 & 0.8505 & 0.7523 & 0.7346 & 0.5475 \\
    Retain       & 0.8058 & 0.8498 & 0.7470 & 0.7319 & 0.5287 \\
    Adacare      & 0.7650 & 0.8140 & 0.7050 & 0.6866 & 0.5744 \\
    ConCare      & 0.7467 & 0.8014 & 0.7070 & 0.6756 & 0.6029 \\
    \bottomrule
  \end{tabular}
\end{table}

\FloatBarrier
\subsection{Clinical Imaging}
\label{app:image}

\begin{table}[!ht]
  \centering
  \caption{Chest X-ray classification on Covid19CXR~\cite{cohen2020covid19cxr}.
  ROC-AUC is computed one-vs-rest (OvR).}
  \label{tab:covid-cxr}
  \begin{tabular}{lccc}
    \toprule
    \textbf{Model} & \textbf{Accuracy} & \textbf{F1 Macro} & \textbf{ROC-AUC (OvR)} \\
    \midrule
    CNN        & 0.8649 & 0.8741 & 0.9750 \\
    ResNet-18  & 0.9532 & 0.9601 & 0.9942 \\
    ViT-B/32   & 0.8880 & 0.8926 & 0.9773 \\
    \bottomrule
  \end{tabular}
\end{table}

\FloatBarrier
\subsection{Clinical Text}
\label{app:text}

\begin{table}[!ht]
  \centering
  \caption{Medical Transcriptions classification.}
  \label{tab:med-transcriptions}
  \begin{tabular}{lcc}
    \toprule
    \textbf{Model} & \textbf{Accuracy} & \textbf{F1 Macro} \\
    \midrule
    Text Embedding              & 0.3400 & 0.0664 \\
    Text Embedding (BERT Base)  & 0.3622 & 0.0470 \\
    Text Embedding (BioBERT)    & 0.3702 & 0.0718 \\
    \bottomrule
  \end{tabular}
\end{table}

\FloatBarrier
\subsection{Biosignals (EEG)}
\label{app:eeg}

\begin{table}[!ht]
  \centering
  \caption{Abnormal EEG detection on the TUH Abnormal EEG Corpus
  (TUAB)~\cite{obeid2016tuab}.}
  \label{tab:tuab}
  \begin{tabular}{lccc}
    \toprule
    \textbf{Model} & \textbf{ROC-AUC} & \textbf{Accuracy} & \textbf{F1 Macro} \\
    \midrule
    BIOT           & 0.8923 & 0.8099 & 0.7695 \\
    ContraWR       & 0.8334 & 0.7619 & 0.7278 \\
    SPaRCNet       & 0.8552 & 0.7677 & 0.6996 \\
    TFM-Tokenizer  & 0.8903 & 0.8113 & 0.7783 \\
    \bottomrule
  \end{tabular}
\end{table}

\begin{table}[!ht]
  \centering
  \caption{Event classification on the TUH EEG Events Corpus
  (TUEV)~\cite{shah2018tuev}.}
  \label{tab:tuev}
  \begin{tabular}{lcc}
    \toprule
    \textbf{Model} & \textbf{Accuracy} & \textbf{F1 Macro} \\
    \midrule
    BIOT           & 0.7162 & 0.4218 \\
    ContraWR       & 0.7411 & 0.5026 \\
    SPaRCNet       & 0.6909 & 0.3642 \\
    TFM-Tokenizer  & 0.7768 & 0.5451 \\
    \bottomrule
  \end{tabular}
\end{table}

\FloatBarrier
\subsection{Non-EHR Data-Processing Benchmarks}
\label{app:processing}

Table~\ref{tab:processing-benchmarks} reports the data-processing cost
(dataset loading, task processing, and peak memory) for the imaging, text,
and biosignal datasets, complementing the EHR processing benchmarks in the
main paper.

\begin{table}[!ht]
  \centering
  \caption{Data-processing benchmarks for non-EHR modalities:
  Covid19CXR~\cite{cohen2020covid19cxr}, Medical Transcriptions,
  TUAB~\cite{obeid2016tuab}, and TUEV~\cite{shah2018tuev}. Wall time is
  the sum of dataset loading and task processing.}
  \label{tab:processing-benchmarks}
  \begin{tabular}{llccc}
    \toprule
    \textbf{Modality} & \textbf{Dataset} & \textbf{Loading (s)} & \textbf{Task Processing (s)} & \textbf{Peak Mem.\ (GB)} \\
    \midrule
    Image          & Covid19CXR             & 11.74 & 131.05  & 1.95 \\
    Text           & Medical Transcriptions & 16.73 & 31.73   & 2.24 \\
    Signals (EEG)  & TUAB                   & 0.60  & 6547.18 & 2.05 \\
    Signals (EEG)  & TUEV                   & 0.19  & 1210.60 & 2.78 \\
    \bottomrule
  \end{tabular}
\end{table}

\FloatBarrier
\subsection{Interpretability}
\label{app:interpretability}

We additionally benchmark post-hoc interpretability methods on a Transformer
model trained for mortality prediction, reported using the comprehensiveness
and sufficiency faithfulness metrics. Best values per metric are shown in
\textbf{bold} (higher comprehensiveness and lower sufficiency are better).

\begin{table}[!ht]
  \centering
  \caption{Interpretability benchmarks for mortality prediction on
  MIMIC-IV~\cite{johnson2023mimic4} with a Transformer model. Higher
  comprehensiveness and lower sufficiency indicate more faithful
  attributions.}
  \label{tab:interpretability}
  \begin{tabular}{lcc}
    \toprule
    \textbf{Method} & \textbf{Comprehensiveness} & \textbf{Sufficiency} \\
    \midrule
    Integrated Gradient & 0.6013 & $0.0290$ \\
    DeepLIFT            & 0.3228 & $0.1573$ \\
    GIM                 & 0.5712 & $0.0392$ \\
    SHAP                & 0.4800 & $0.1163$ \\
    LIME                & 0.5220 & $0.1477$ \\
    Attn-Grad           & \textbf{0.6031} & $\mathbf{-0.0081}$ \\
    \bottomrule
  \end{tabular}
\end{table}

\FloatBarrier
\subsection{Uncertainty Quantification}
\label{app:uq}

Finally, we benchmark conformal prediction methods on TUEV at a target
miscoverage level of $\alpha = 0.1$ (i.e., a target coverage of $0.9$).
Results are averaged over random sample splits and reported as
mean~$\pm$~standard deviation.

\begin{table}[!ht]
  \centering
  \caption{Conformal prediction on TUEV~\cite{shah2018tuev} ($\alpha = 0.1$,
  random sample splits). Empirical coverage and prediction set size are
  reported as mean~$\pm$~standard deviation.}
  \label{tab:conformal-prediction}
  \begin{tabular}{lcc}
    \toprule
    \textbf{Method} & \textbf{Empirical Coverage} & \textbf{Set Size} \\
    \midrule
    Conformal Prediction              & $0.7461 \pm 0.0334$ & $1.11 \pm 0.21$ \\
    \quad + KDE Covariate Shift Adj.  & $0.7457 \pm 0.0072$ & $1.11 \pm 0.14$ \\
    \quad + KMeans Adjustment         & $0.7488 \pm 0.0152$ & $1.23 \pm 0.24$ \\
    Neighborhood Conformal Prediction & $0.9152 \pm 0.0117$ & $1.25 \pm 0.13$ \\
    \bottomrule
  \end{tabular}
\end{table}

\FloatBarrier


\clearpage
\section{PyHealth Datasets} \label{appdx: datasets}

PyHealth provides a comprehensive collection of healthcare datasets spanning multiple modalities and clinical domains. These datasets enable researchers to develop and evaluate machine learning models across diverse healthcare applications, from electronic health records to medical imaging, physiological signals, genomics, and clinical text. Each dataset is pre-processed and standardized to facilitate seamless integration with PyHealth's modeling and task frameworks.
\begin{figure*}[h]
    \centering
    \includegraphics[width=0.95\textwidth]{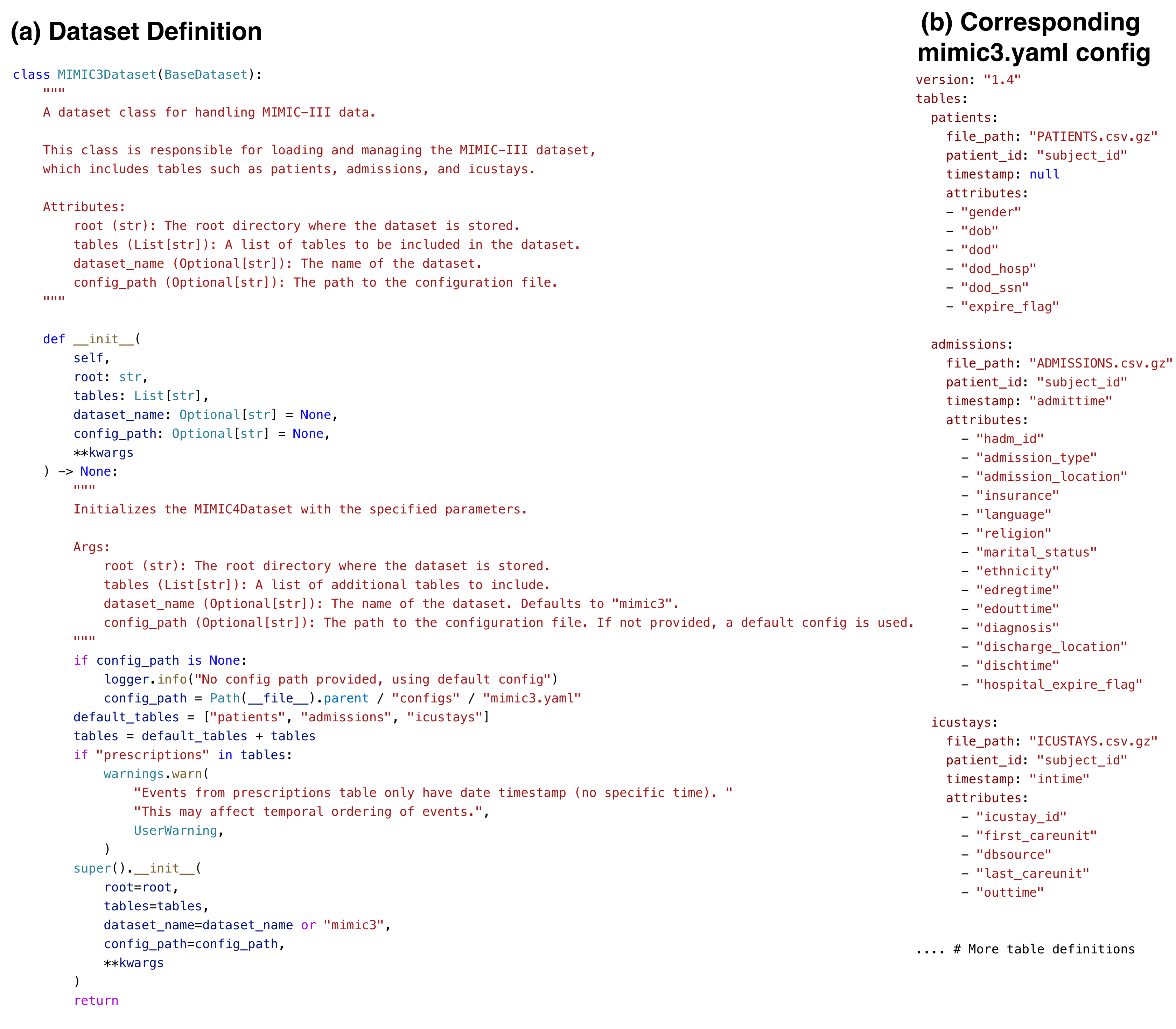}
    \caption{\textbf{PyHealth 2.0 Dataset Example.}  Defining a PyHealth dataset simply requires the inheritance of the (a) BaseDataset class and the definition of a (b) mimic3.yaml file for direct deployment.}
    \label{fig: dataset example}
   \vspace{-0.3cm}
\end{figure*}

\begin{table}[htbp]
\centering
\caption{Overview of PyHealth Dataset Categories}
\label{tab:dataset_overview}
\begin{tabularx}{\textwidth}{lcp{0.6\textwidth}}
\toprule
\textbf{Category} & \textbf{Count} & \textbf{Description} \\
\midrule
EHR Datasets & 7 & Electronic health records from ICU and hospital settings, including MIMIC-III/IV, eICU, and standardized OMOP formats \\
\midrule
Medical Imaging & 3 & Chest X-ray datasets for disease classification and COVID-19 detection \\
\midrule
Physiological Signals & 8 & EEG, ECG, and heart sound recordings for sleep staging, seizure detection, and cardiac monitoring \\
\midrule
Genomics & 3 & Genetic variant and cancer mutation databases for pathogenicity prediction and survival analysis \\
\midrule
Clinical Text & 1 & Medical transcription dataset for specialty classification \\
\bottomrule
\end{tabularx}
\end{table}

\subsection*{EHR Datasets}

Electronic health record (EHR) datasets form the foundation of healthcare AI research, providing rich longitudinal patient data including diagnoses, procedures, medications, laboratory results, and clinical notes. PyHealth supports major publicly available EHR databases from intensive care units and broader hospital settings, along with standardized formats like OMOP that enable cross-institutional research.

\begin{table*}[htbp]
\centering
\caption{EHR Datasets in PyHealth}
\label{tab:ehr_datasets}
\begin{tabularx}{\textwidth}{l Y l Y}
\toprule
\textbf{Dataset} & \textbf{Description} & \textbf{Data Type} & \textbf{Key Features} \\
\midrule
MIMIC3Dataset & MIMIC-III Clinical Database v1.4 \cite{johnson2016mimiciii} & EHR (ICU) & 46K+ patients, diagnoses, procedures, prescriptions, lab events, clinical notes \\
\midrule
MIMIC4Dataset & MIMIC-IV Clinical Database v2.0+ \cite{johnson2023mimic4} & EHR (Hospital + ICU) & 300K+ patients, improved de-identification, expanded tables \\
\midrule
eICUDataset & eICU Collaborative Research Database \cite{pollard2018eicu} & EHR (ICU) & 200K+ ICU stays, 335 hospitals, diagnoses, medications, treatments \\
\midrule
OMOPDataset & OMOP Common Data Model format \cite{hripcsak2015omop} & EHR (Standardized) & Supports any OMOP-formatted data \\
\midrule
MIMICExtractDataset & MIMIC-Extract preprocessed benchmark \cite{wang2020mimicextract} & EHR (Preprocessed) & Pre-processed features, standardized cohorts \\
\midrule
EHRShotDataset & EHRShot benchmark dataset \cite{wornow2023ehrshot} & EHR (Benchmark) & 15 predictive tasks \\
\midrule
Support2Dataset & SUPPORT Study II survival data \cite{knaus1995support} & EHR (Survival) & 9,105 patients, survival outcomes \\
\bottomrule
\end{tabularx}
\end{table*}

\subsection*{Medical Imaging Datasets}

Medical imaging datasets enable the development of computer vision models for diagnostic support. PyHealth provides access to large-scale chest X-ray collections annotated with disease labels, supporting both single-label and multi-label classification tasks for thoracic pathologies and infectious diseases.

\begin{table*}[htbp]
\centering
\caption{Medical Imaging Datasets in PyHealth}
\label{tab:imaging_datasets}
\begin{tabularx}{\textwidth}{l Y l Y}
\toprule
\textbf{Dataset} & \textbf{Description} & \textbf{Data Type} & \textbf{Key Features} \\
\midrule
ChestXray14Dataset & NIH ChestX-ray14 Dataset \cite{wang2017chestxray14} & Chest X-ray & 112K images, 14 disease labels \\
\midrule
COVID19CXRDataset & COVID-19 Chest X-ray Dataset \cite{cohen2020covid19cxr} & Chest X-ray & COVID-19/Normal/Pneumonia classification \\
\midrule
MIMICCXRDataset & MIMIC-CXR Chest X-ray Database \cite{johnson2019mimic_cxr} & Chest X-ray & 377K images, linked to MIMIC-IV \\
\bottomrule
\end{tabularx}
\end{table*}

\subsection*{Physiological Signal Datasets}

Physiological signal datasets capture time-series recordings of brain activity (EEG), cardiac rhythms (ECG), and heart sounds (PCG). These datasets support the development of deep learning models for automated sleep staging, seizure detection, cardiac arrhythmia classification, and other diagnostic tasks that traditionally require expert manual annotation.

\begin{table*}[htbp]
\centering
\caption{Physiological Signal Datasets in PyHealth}
\label{tab:signal_datasets}
\begin{tabularx}{\textwidth}{l Y l Y}
\toprule
\textbf{Dataset} & \textbf{Description} & \textbf{Data Type} & \textbf{Key Features} \\
\midrule
SleepEDFDataset & Sleep-EDF Database Expanded \cite{kemp2000sleepedf} & EEG (Sleep) & 197 recordings, 5 sleep stages \\
\midrule
ISRUCDataset & ISRUC-Sleep Dataset \cite{khalighi2016isruc} & EEG (Sleep) & 100 subjects, 6-channel EEG, dual expert annotations \\
\midrule
SHHSDataset & Sleep Heart Health Study \cite{quan1997shhs} & EEG (Sleep) & 6K+ recordings, cardiovascular outcomes \\
\midrule
TUABDataset & TUH Abnormal EEG Corpus \cite{obeid2016tuab} & EEG (Clinical) & Abnormal vs normal EEG classification \\
\midrule
TUEVDataset & TUH EEG Events Corpus \cite{shah2018tuev} & EEG (Events) & EEG event detection, 6 event classes \\
\midrule
CardiologyDataset \cite{PhysioNet-challenge-2020-1.0.2_Cardiology} & Cardiology ECG Dataset & ECG & Multiple arrhythmia types, 12-lead ECG \\
\midrule
BMDHSDataset  \cite{ali2024buetmultidiseaseheartsound_bmd_hs} & BMD Heart Sound Dataset & PCG (Audio) & Heart valve disease, 8 recordings/patient \\
\midrule
DREAMTDataset \cite{PhysioNet-dreamt-2.1.0} & DREAMT Sleep Dataset & EEG (Sleep) & Multi-channel polysomnography \\
\bottomrule
\end{tabularx}
\end{table*}

\subsection*{Genomics Datasets}
Genomics datasets provide genetic and molecular information for precision medicine applications. These include variant databases with clinical significance annotations, somatic mutation catalogs from cancer patients, and multi-omics datasets linking genomic profiles to patient outcomes. Such datasets enable the development of models for pathogenicity prediction, mutation burden estimation, and survival analysis. From our team of PyHealth researchers, we contribute 3 different benchmarks that compose of our prostate cancer variant benchmark \citep{tavara2025prostate_varbench}.

\begin{table*}[htbp]
\centering
\caption{Genomics Datasets in PyHealth}
\label{tab:genomics_datasets}
\begin{tabularx}{\textwidth}{l Y l Y}
\toprule
\textbf{Dataset} & \textbf{Description} & \textbf{Data Type} & \textbf{Key Features} \\
\midrule
ClinVarDataset & ClinVar Genetic Variant Database \cite{landrum2018clinvar} & Variants & Clinical significance classification \\
\midrule
COSMICDataset & COSMIC Cancer Mutation Database \cite{tate2019cosmic} & Mutations & Somatic mutations, FATHMM pathogenicity \\
\midrule
TCGAPRADDataset & TCGA Prostate Adenocarcinoma \cite{tcga2015prad} & Multi-omics & Mutations, clinical data, survival \\
\bottomrule
\end{tabularx}
\end{table*}

\subsection*{Clinical Text Dataset}

Clinical text datasets contain unstructured medical narratives such as discharge summaries, radiology reports, and operative notes. These enable natural language processing research for medical specialty classification, named entity recognition, and clinical information extraction tasks.

\begin{table*}[htbp]
\centering
\caption{Clinical Text Dataset in PyHealth}
\label{tab:text_datasets}
\begin{tabularx}{\textwidth}{l Y l Y}
\toprule
\textbf{Dataset} & \textbf{Description} & \textbf{Data Type} & \textbf{Key Features} \\
\midrule
MedicalTranscriptionsDataset & MTSamples Medical Transcriptions & Text & Medical specialty classification \\
\bottomrule
\end{tabularx}
\end{table*}

\clearpage
\section{PyHealth Tasks} \label{appdx: tasks}

PyHealth provides a diverse collection of pre-defined clinical prediction tasks that transform raw healthcare data into machine learning-ready formats. These tasks implement standardized data preprocessing pipelines, feature extraction, and label generation for common healthcare prediction objectives. Each task is designed to work seamlessly with specific datasets and supports various model architectures through a unified interface.

\begin{table}[htbp]
\centering
\caption{Overview of PyHealth Task Categories}
\label{tab:task_overview}
\begin{tabularx}{\textwidth}{lcp{0.6\textwidth}}
\toprule
\textbf{Task Category} & \textbf{Count} & \textbf{Description} \\
\midrule
Mortality Prediction & 9 & In-hospital and ICU mortality prediction across multiple EHR databases \\
\midrule
Readmission Prediction & 4 & 30-day and general readmission risk prediction \\
\midrule
Drug Recommendation & 2 & Medication combination recommendation with safety constraints \\
\midrule
Length of Stay & 4 & Hospital length of stay classification into discrete bins \\
\midrule
Sleep Staging & 4 & Automated sleep stage classification from EEG signals \\
\midrule
EEG Analysis & 2 & Abnormality detection and event classification \\
\midrule
Medical Imaging & 3 & Disease classification from chest X-rays \\
\midrule
Cardiology Detection & 5 & ECG-based cardiac condition identification \\
\midrule
Genomics/Cancer & 4 & Variant pathogenicity and cancer survival prediction \\
\midrule
Other Specialized & 6 & Benchmark suites, medical coding, and patient linkage \\
\bottomrule
\end{tabularx}
\end{table}

\subsection*{Mortality Prediction Tasks}

Mortality prediction tasks aim to identify patients at high risk of death during hospitalization or within a specified timeframe. These tasks are critical for clinical decision support, resource allocation, and early intervention. PyHealth supports mortality prediction across multiple EHR databases with varying feature representations, including structured codes, laboratory values, and multimodal data incorporating clinical notes and medical images.

\begin{table*}[htbp]
\centering
\caption{Mortality Prediction Tasks in PyHealth}
\label{tab:mortality_tasks}
\begin{tabularx}{\textwidth}{Y l l Y}
\toprule
\textbf{Task} & \textbf{Dataset} & \textbf{Output} & \textbf{Input Features} \\
\midrule
MortalityPredictionMIMIC3 & MIMIC-III & Binary & conditions, procedures, drugs \\
\midrule
MortalityPredictionMIMIC4 & MIMIC-IV & Binary & conditions, procedures, drugs \\
\midrule
MultimodalMortalityPrediction-MIMIC3 & MIMIC-III & Binary & conditions, procedures, drugs, clinical\_notes \\
\midrule
MultimodalMortalityPrediction-MIMIC4 & MIMIC-IV & Binary & conditions, procedures, drugs, discharge, radiology, labs, image \\
\midrule
MortalityPredictionEICU & eICU & Binary & conditions, procedures, drugs \\
\midrule
MortalityPredictionEICU2 & eICU & Binary & conditions (admissionDx), procedures (treatment) \\
\midrule
MortalityPredictionOMOP & OMOP & Binary & conditions, procedures, drugs \\
\midrule
MortalityPrediction-StageNetMIMIC4 & MIMIC-IV & Binary & icd\_codes (stagenet), labs (tensor) \\
\midrule
InHospitalMortalityMIMIC4 & MIMIC-IV & Binary & labs (timeseries) \\
\bottomrule
\end{tabularx}
\end{table*}

\subsection*{Readmission Prediction Tasks}

Hospital readmission prediction identifies patients likely to return to the hospital within a specified period (commonly 30 days) after discharge. Reducing preventable readmissions improves patient outcomes and reduces healthcare costs. These tasks leverage comprehensive patient histories including diagnoses, procedures, and medications to assess readmission risk.

\begin{table*}[htbp]
\centering
\caption{Readmission Prediction Tasks in PyHealth}
\label{tab:readmission_tasks}
\begin{tabularx}{\textwidth}{Y l l Y}
\toprule
\textbf{Task} & \textbf{Dataset} & \textbf{Output} & \textbf{Input Features} \\
\midrule
ReadmissionPredictionMIMIC3 & MIMIC-III & Binary & conditions, procedures, drugs \\
\midrule
Readmission30DaysMIMIC4 & MIMIC-IV & Binary & conditions, procedures, drugs \\
\midrule
readmission\_prediction\_eicu\_fn & eICU & Binary & conditions, procedures, drugs \\
\midrule
ReadmissionPredictionOMOP & OMOP & Binary & conditions, procedures, drugs \\
\bottomrule
\end{tabularx}
\end{table*}

\subsection*{Drug Recommendation Tasks}

Drug recommendation tasks predict optimal medication combinations for patients based on their medical history and current conditions. These tasks are particularly challenging due to the need to avoid harmful drug-drug interactions (DDI) while maximizing therapeutic efficacy. The output is typically a multilabel prediction where multiple medications can be recommended simultaneously.

\begin{table*}[htbp]
\centering
\caption{Drug Recommendation Tasks in PyHealth}
\label{tab:drug_tasks}
\begin{tabularx}{\textwidth}{Y l l Y}
\toprule
\textbf{Task} & \textbf{Dataset} & \textbf{Output} & \textbf{Input Features} \\
\midrule
DrugRecommendationMIMIC3 & MIMIC-III & Multilabel & conditions, procedures, drugs\_hist (nested) \\
\midrule
DrugRecommendationMIMIC4 & MIMIC-IV & Multilabel & conditions, procedures, drugs\_hist (nested) \\
\bottomrule
\end{tabularx}
\end{table*}

\subsection*{Length of Stay Prediction Tasks}

Length of stay (LOS) prediction estimates how long a patient will remain hospitalized, typically discretized into clinically meaningful bins (e.g., 0-1 days, 1-2 days, etc.). Accurate LOS prediction supports hospital resource planning, bed management, and discharge planning. These tasks frame the problem as multiclass classification with 10 ordinal categories.

\begin{table*}[htbp]
\centering
\caption{Length of Stay Prediction Tasks in PyHealth}
\label{tab:los_tasks}
\begin{tabularx}{\textwidth}{Y l l Y}
\toprule
\textbf{Task} & \textbf{Dataset} & \textbf{Output} & \textbf{Input Features} \\
\midrule
LengthOfStayPredictionMIMIC3 & MIMIC-III & Multiclass (10) & conditions, procedures, drugs \\
\midrule
LengthOfStayPredictionMIMIC4 & MIMIC-IV & Multiclass (10) & conditions, procedures, drugs \\
\midrule
LengthOfStayPredictioneICU & eICU & Multiclass (10) & conditions, procedures, drugs \\
\midrule
LengthOfStayPredictionOMOP & OMOP & Multiclass (10) & conditions, procedures, drugs \\
\bottomrule
\end{tabularx}
\end{table*}

\subsection*{Sleep Staging Tasks}

Sleep staging tasks classify EEG recordings into distinct sleep stages (Wake, N1, N2, N3, REM) following standard polysomnography scoring criteria. Automated sleep staging reduces the burden of manual annotation by sleep technologists and enables large-scale sleep research. These tasks process multi-channel EEG signals segmented into 30-second epochs.

\begin{table*}[htbp]
\centering
\caption{Sleep Staging Tasks in PyHealth}
\label{tab:sleep_tasks}
\begin{tabularx}{\textwidth}{Y l l Y}
\toprule
\textbf{Task} & \textbf{Dataset} & \textbf{Output} & \textbf{Input Features} \\
\midrule
sleep\_staging\_sleepedf\_fn & SleepEDF & Multiclass (5-6) & EEG signal epochs \\
\midrule
sleep\_staging\_isruc\_fn & ISRUC & Multiclass (5) & EEG signal epochs \\
\midrule
sleep\_staging\_shhs\_fn & SHHS & Multiclass (5) & EEG signal epochs \\
\midrule
SleepStagingSleepEDF & SleepEDF & Multiclass & EEG signal (class-based) \\
\bottomrule
\end{tabularx}
\end{table*}

\subsection*{EEG Analysis Tasks}

EEG analysis tasks focus on detecting abnormalities and specific events in electroencephalography recordings. Abnormality detection classifies entire EEG recordings as normal or abnormal, supporting clinical screening workflows. Event detection identifies transient patterns such as seizures, spikes, and other neurological phenomena that require medical attention.

\begin{table*}[htbp]
\centering
\caption{EEG Analysis Tasks in PyHealth}
\label{tab:eeg_tasks}
\begin{tabularx}{\textwidth}{Y l l Y}
\toprule
\textbf{Task} & \textbf{Dataset} & \textbf{Output} & \textbf{Input Features} \\
\midrule
EEG\_isAbnormal\_fn & TUAB & Binary & 16-channel bipolar EEG \\
\midrule
EEGEventsTUEV & TUEV & Multiclass (6) & 16-channel bipolar EEG \\
\bottomrule
\end{tabularx}
\end{table*}

\subsection*{Medical Imaging Tasks}

Medical imaging tasks apply computer vision techniques to diagnostic images, particularly chest X-rays. These tasks range from binary classification of single pathologies to multilabel classification where multiple conditions may be present simultaneously. They support radiologist workflow augmentation and automated screening in resource-limited settings.

\begin{table*}[htbp]
\centering
\caption{Medical Imaging Tasks in PyHealth}
\label{tab:imaging_tasks}
\begin{tabularx}{\textwidth}{Y l l Y}
\toprule
\textbf{Task} & \textbf{Dataset} & \textbf{Output} & \textbf{Input Features} \\
\midrule
ChestXray14BinaryClassification & ChestXray14 & Binary & X-ray image \\
\midrule
ChestXray14MultilabelClassification & ChestXray14 & Multilabel (14) & X-ray image \\
\midrule
COVID19CXRClassification & COVID19CXR & Multiclass (3) & X-ray image \\
\bottomrule
\end{tabularx}
\end{table*}

\subsection*{Cardiology Detection Tasks}

Cardiology detection tasks identify specific cardiac conditions from ECG recordings. Each task targets a clinically relevant condition such as valve disease, conduction disorders, or arrhythmias. These binary classification tasks enable automated ECG interpretation and can serve as diagnostic support tools or screening mechanisms in primary care settings.

\begin{table*}[htbp]
\centering
\caption{Cardiology Detection Tasks in PyHealth}
\label{tab:cardiology_tasks}
\begin{tabularx}{\textwidth}{Y l l Y}
\toprule
\textbf{Task} & \textbf{Dataset} & \textbf{Output} & \textbf{Condition Detected} \\
\midrule
cardiology\_isAR\_fn & Cardiology & Binary & Aortic Regurgitation \\
\midrule
cardiology\_isBBBFB\_fn & Cardiology & Binary & Bundle Branch Block / Fascicular Block \\
\midrule
cardiology\_isAD\_fn & Cardiology & Binary & Atrial Disorders \\
\midrule
cardiology\_isCD\_fn & Cardiology & Binary & Conduction Disorders \\
\midrule
cardiology\_isWA\_fn & Cardiology & Binary & Wolf-Parkinson-White / Arrhythmias \\
\bottomrule
\end{tabularx}
\end{table*}

\subsection*{Genomics and Cancer Tasks}

Genomics and cancer tasks predict clinically relevant outcomes from genetic and molecular data. These include classifying the pathogenicity of genetic variants, estimating tumor mutation burden, and predicting patient survival based on genomic profiles. Such tasks advance precision oncology by enabling risk stratification and treatment selection based on molecular characteristics.

\begin{table*}[htbp]
\centering
\caption{Genomics and Cancer Tasks in PyHealth}
\label{tab:genomics_tasks}
\begin{tabularx}{\textwidth}{Y l l Y}
\toprule
\textbf{Task} & \textbf{Dataset} & \textbf{Output} & \textbf{Input Features} \\
\midrule
CancerSurvivalPrediction & TCGA-PRAD & Binary & mutations, age, Gleason score \\
\midrule
CancerMutationBurden & TCGA-PRAD & Binary & mutations, age \\
\midrule
VariantClassificationClinVar & ClinVar & Multiclass (5) & gene\_symbol, variant\_type, chromosome \\
\midrule
MutationPathogenicityPrediction & COSMIC & Binary & gene\_name, mutation\_description, primary\_site \\
\bottomrule
\end{tabularx}
\end{table*}

\subsection*{Other Specialized Tasks}

Additional specialized tasks address diverse healthcare applications including benchmark evaluations, medical coding from clinical notes, patient record linkage, and heart sound classification. These tasks demonstrate PyHealth's versatility across different healthcare informatics problems beyond traditional clinical prediction.

\begin{table*}[htbp]
\centering
\caption{Other Specialized Tasks in PyHealth}
\label{tab:other_tasks}
\begin{tabularx}{\textwidth}{Y l l Y}
\toprule
\textbf{Task} & \textbf{Dataset} & \textbf{Output} & \textbf{Description} \\
\midrule
BenchmarkEHRShot & EHRShot & Various & 15 benchmark tasks \\
\midrule
BMDHSDiseaseClassification & BMD-HS & Multilabel (4) & Heart valve disease (AS, AR, MR, MS) \\
\midrule
SurvivalPreprocessSupport2 & Support2 & Survival & Survival outcome preprocessing \\
\midrule
MedicalTranscriptionsClassification & MTSamples & Multiclass & Medical specialty from transcription \\
\midrule
MIMIC3ICD9Coding & MIMIC-III & Multilabel & ICD-9 code prediction from notes \\
\midrule
patient\_linkage\_mimic3\_fn & MIMIC-III & Linkage & Patient record linkage \\
\bottomrule
\end{tabularx}
\end{table*}

\clearpage
\section{PyHealth Models} \label{appdx: models}

PyHealth implements a comprehensive collection of machine learning and deep learning models tailored for healthcare applications. The model library spans general-purpose sequence models adapted for medical data, healthcare-specific architectures designed for clinical prediction tasks, specialized drug recommendation models that incorporate pharmacological knowledge, and auxiliary models for generation, graph learning, and transfer learning. All models follow a unified API that enables consistent training, evaluation, and deployment workflows.
\begin{figure*}[h]
    \centering
    \includegraphics[width=1.0\textwidth]{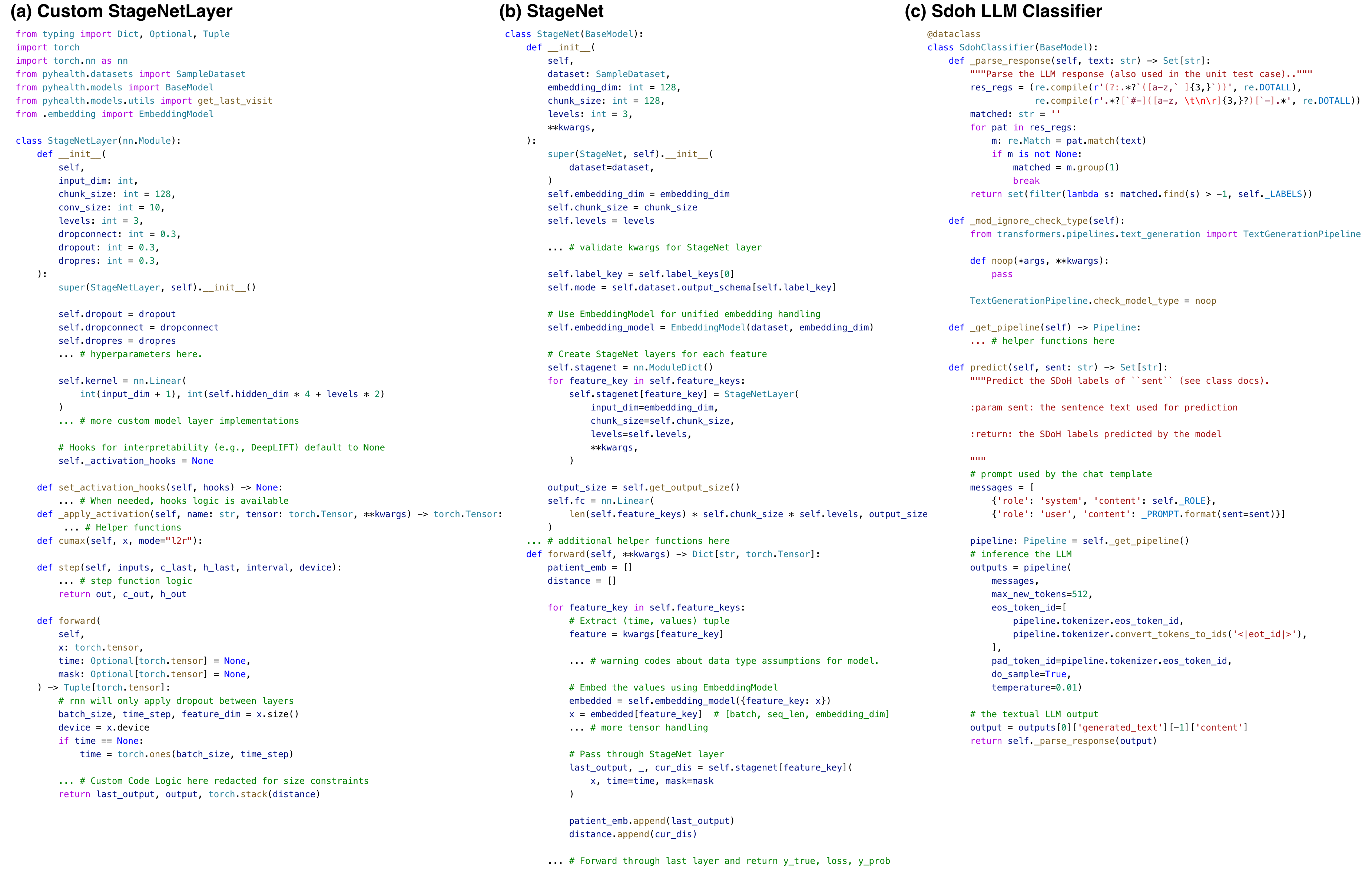}
    \caption{\textbf{PyHealth 2.0 Models.}  PyHealth offers examples of modular model layers like the StageNetLayer (a) as well as custom models like StageNet (b), and even LLMs for social determinants of health extraction (c) \citep{guevara2024sdoh, landes2025integrationlargelanguagemodels_sdoh}. In this case, PyHealth's model class makes very little assumptions on what's allowed for model usage beyond requirements in documentation as shown by the LLM wrapper (c). However, for direct use with pre-existing dataloaders and the built-in trainer, we highly recommend users to follow a more strict forward() operation where users are recommended to return the predictions and loss functions of their models (b) and more closely follow schema requirements made from PyHealth tasks here. }
    \label{fig: modeling example}
   \vspace{-0.3cm}
\end{figure*}

\begin{table}[htbp]
\centering
\caption{Overview of PyHealth Model Categories}
\label{tab:model_overview}
\begin{tabularx}{\textwidth}{lcp{0.6\textwidth}}
\toprule
\textbf{Model Category} & \textbf{Count} & \textbf{Description} \\
\midrule
General Sequence Models & 5 & RNN, Transformer, CNN, TCN, and MLP architectures for general healthcare prediction \\
\midrule
Healthcare-Specific & 10 & Models designed specifically for EHR and clinical data including RETAIN, StageNet, AdaCare, and ConCare \\
\midrule
Drug Recommendation & 4 & Specialized models for safe medication combination recommendation with DDI constraints \\
\midrule
Specialized Models & 9 & Graph neural networks, generative models, LLM-based classifiers, and pre-trained vision/language models \\
\bottomrule
\end{tabularx}
\end{table}

\subsection*{General Sequence Models}

General sequence models provide foundational architectures for processing temporal healthcare data. These models can be applied across different prediction tasks and data modalities. Recurrent neural networks (RNNs) capture temporal dependencies in patient visits, Transformers leverage attention mechanisms for long-range interactions, convolutional neural networks (CNNs) extract local patterns from signals and images, temporal convolutional networks (TCNs) offer efficient sequence modeling, and multi-layer perceptrons (MLPs) serve as baseline classifiers.

\begin{table*}[htbp]
\centering
\caption{General Sequence Models in PyHealth}
\label{tab:general_models}
\begin{tabularx}{\textwidth}{l Y l Y}
\toprule
\textbf{Model} & \textbf{Architecture} & \textbf{Task Types} & \textbf{Reference} \\
\midrule
RNN & Recurrent Neural Network (GRU/LSTM/Vanilla) & Classification, Regression & \cite{hochreiter1997lstm,cho2014gru} \\
\midrule
Transformer & Self-attention based encoder & Classification, Regression & \cite{vaswani2017transformer} \\
\midrule
CNN & Convolutional Neural Network & Signal/Image tasks & \cite{lecun1998cnn} \\
\midrule
TCN & Temporal Convolutional Network & Sequence tasks & \cite{bai2018tcn} \\
\midrule
\cite{rumelhart1986learning}\\
\bottomrule
\end{tabularx}
\end{table*}

\subsection*{Healthcare-Specific Models}

Healthcare-specific models incorporate domain knowledge and specialized architectures designed for clinical data characteristics. RETAIN provides interpretable predictions through reverse-time attention, StageNet captures disease progression stages, AdaCare and ConCare adapt feature importance based on patient context, GRASP leverages patient similarity graphs, Agent employs multi-agent reinforcement learning, Deepr applies CNNs to medical record sequences, and SparcNet and ContraWR handle physiological signals with sparse convolutions and contrastive learning respectively.

\begin{table*}[htbp]
\centering
\caption{Healthcare-Specific Models in PyHealth}
\label{tab:healthcare_models}
\begin{tabularx}{\textwidth}{l Y l Y}
\toprule
\textbf{Model} & \textbf{Description} & \textbf{Task Types} & \textbf{Reference} \\
\midrule
RETAIN & Reverse Time Attention Model with interpretable attention & Prediction & \cite{choi2016retain} \\
\midrule
StageNet & Stage-aware neural network for health risk prediction & Prediction & \cite{gao2020stagenet} \\
\midrule
StageAttentionNet & StageNet with Multi-Head Attention between SA-LSTM and CNN & Prediction & Extended StageNet \\
\midrule
AdaCare & Adaptive feature importance calibration for EHR & Prediction & \cite{ma2020adacare} \\
\midrule
ConCare & Context-aware health status representation learning & Prediction & \cite{ma2020concare} \\
\midrule
GRASP & Graph-based patient similarity with clustering & Prediction & \cite{zhang2021grasp} \\
\midrule
Agent & Dr. Agent - Multi-agent RL with dynamic skip connections & Prediction & \cite{gao2020agent} \\
\midrule
Deepr & CNN for medical records with max pooling & Prediction & \cite{nguyen2017deepr} \\
\midrule
SparcNet & Sparse CNN for EEG signals & Signal & \cite{jing2023sparcnet} \\
\midrule
ContraWR & Contrastive learning for waveform recognition & Signal & \cite{yang2023contrawr} \\
\bottomrule
\end{tabularx}
\end{table*}

\subsection*{Drug Recommendation Models}

Drug recommendation models address the complex problem of suggesting safe and effective medication combinations. SafeDrug incorporates drug-drug interaction (DDI) knowledge graphs with dual molecular encoders, GAMENet uses graph-augmented memory networks to capture medication dependencies, MICRON models medication changes with recurrent residual connections, and MoleRec leverages molecular substructure representations for improved generalization across drug vocabularies.

\begin{table*}[htbp]
\centering
\caption{Drug Recommendation Models in PyHealth}
\label{tab:drug_models}
\begin{tabularx}{\textwidth}{l Y l Y}
\toprule
\textbf{Model} & \textbf{Description} & \textbf{Task Type} & \textbf{Reference} \\
\midrule
SafeDrug & Safe drug recommendation with DDI knowledge graph & Drug Rec. & \cite{yang2021safedrug} \\
\midrule
GAMENet & Graph Augmented Memory Network for drug recommendation & Drug Rec. & \cite{shang2019gamenet} \\
\midrule
MICRON & Change-aware drug recommendation with DDI & Drug Rec. & \cite{yang2021micron} \\
\midrule
MoleRec & Molecular structure-aware drug recommendation & Drug Rec. & \cite{yang2023molerec} \\
\bottomrule
\end{tabularx}
\end{table*}

\subsection*{Specialized Models}

Specialized models extend PyHealth's capabilities to additional domains and methodologies. SdohClassifier applies large language models with LoRA fine-tuning for social determinants of health extraction, MedLink performs patient record linkage, VAE and GAN enable generative modeling, GAT and GCN provide graph neural network implementations, TorchvisionModel wraps pre-trained computer vision models, and TransformersModel integrates HuggingFace language models including domain-specific variants like ClinicalBERT.

\begin{table*}[htbp]
\centering
\caption{Specialized Models in PyHealth}
\label{tab:specialized_models}
\begin{tabularx}{\textwidth}{l Y l Y}
\toprule
\textbf{Model} & \textbf{Description} & \textbf{Task Type} & \textbf{Reference} \\
\midrule
SdohClassifier & LLM-based Social Determinants of Health classifier (Llama 3.1 + LoRA) & NLP & \cite{guevara2024sdoh} \\
\midrule
MedLink & Patient linkage across EHR systems & Linkage & Patient record matching \\
\midrule
LogisticRegression & Logistic regression baseline & Classification & Standard baseline \\
\midrule
VAE & Variational Autoencoder & Generation & \cite{kingma2014vae} \\
\midrule
GAN & Generative Adversarial Network & Generation & \cite{goodfellow2014gan} \\
\midrule
GAT & Graph Attention Network & Graph & \cite{velickovic2018gat} \\
\midrule
GCN & Graph Convolutional Network & Graph & \cite{kipf2017gcn} \\
\midrule
TorchvisionModel & Pretrained vision models (ResNet, etc.) & Image & torchvision models \\
\midrule
TransformersModel & HuggingFace Transformers integration & NLP/Multi & \cite{devlin2019bert,alsentzer2019clinicalbert} \\
\bottomrule
\end{tabularx}
\end{table*}

\clearpage

\section{PyHealth Data Processors} \label{appdx: processors}

PyHealth provides a modular system of data processors that transform raw healthcare data into model-ready representations. These processors handle diverse data modalities including sequences, images, signals, text, and structured tabular data. Each processor implements standardized interfaces for fitting vocabularies, encoding features, and batching samples, enabling seamless integration between datasets and models regardless of the underlying data format.

\begin{table}[htbp]
\centering
\caption{Overview of PyHealth Processor Categories}
\label{tab:processor_overview}
\begin{tabularx}{\textwidth}{lcp{0.6\textwidth}}
\toprule
\textbf{Processor Category} & \textbf{Count} & \textbf{Description} \\
\midrule
Sequence Processors & 5 & Handle temporal sequences of medical codes and nested structures \\
\midrule
Signal \& Image & 3 & Process physiological signals, audio, and medical images \\
\midrule
Label Processors & 4 & Encode prediction targets for classification and regression tasks \\
\midrule
Specialized Processors & 5 & Domain-specific processors for StageNet, multi-hot encoding, and raw data \\
\bottomrule
\end{tabularx}
\end{table}

\subsection*{Sequence Processors}

Sequence processors handle temporal medical data including visit sequences, medication histories, and longitudinal patient records. These processors build vocabularies from medical codes (diagnoses, procedures, drugs), perform tokenization, and create padded sequences suitable for recurrent and attention-based models. Nested processors handle hierarchical structures where each visit contains multiple codes, while deep nested processors support additional nesting levels for complex data representations.

\begin{table*}[htbp]
\centering
\caption{Sequence Processors in PyHealth}
\label{tab:sequence_processors}
\begin{tabularx}{\textwidth}{l Y Y}
\toprule
\textbf{Processor} & \textbf{Description} & \textbf{Use Cases} \\
\midrule
SequenceProcessor & Processes flat sequences of tokens/codes & Time series of single values, medication sequences \\
\midrule
NestedSequenceProcessor & Handles sequences of code sets (visits) & EHR visit sequences with multiple codes per visit \\
\midrule
NestedFloatsProcessor & Processes nested sequences with numerical values & Laboratory values, vital signs over visits \\
\midrule
DeepNestedSequenceProcessor & Three-level nesting for complex hierarchies & Drug recommendation with historical medication sets \\
\midrule
DeepNestedFloatsProcessor & Three-level nesting with float values & Multi-visit laboratory panels with multiple tests \\
\bottomrule
\end{tabularx}
\end{table*}

\subsection*{Signal and Image Processors}

Signal and image processors prepare continuous waveforms and medical images for deep learning models. Signal processors handle EEG, ECG, and other physiological time series through resampling, segmentation, and normalization. Image processors load medical images, apply preprocessing transforms, and ensure consistent tensor formatting. Audio processors specifically handle phonocardiogram (heart sound) data with domain-specific preprocessing.

\begin{table*}[htbp]
\centering
\caption{Signal and Image Processors in PyHealth}
\label{tab:signal_image_processors}
\begin{tabularx}{\textwidth}{l Y Y}
\toprule
\textbf{Processor} & \textbf{Description} & \textbf{Use Cases} \\
\midrule
SignalProcessor & Processes 1D continuous signals & EEG, ECG, PPG waveforms \\
\midrule
AudioProcessor & Handles audio waveforms with specialized transforms & Heart sounds, respiratory sounds \\
\midrule
ImageProcessor & Loads and preprocesses medical images & Chest X-rays, CT scans, pathology slides \\
\bottomrule
\end{tabularx}
\end{table*}

\subsection*{Label Processors}

Label processors encode prediction targets into appropriate formats for model training. They handle diverse output types including binary classification, multiclass problems with softmax, multilabel tasks where multiple labels can be active simultaneously, and continuous regression targets. These processors ensure consistent label representations across different tasks and datasets.

\begin{table*}[htbp]
\centering
\caption{Label Processors in PyHealth}
\label{tab:label_processors}
\begin{tabularx}{\textwidth}{l Y Y}
\toprule
\textbf{Processor} & \textbf{Description} & \textbf{Output Format} \\
\midrule
BinaryLabelProcessor & Binary classification labels & Single probability \\
\midrule
MultiClassLabelProcessor & Mutually exclusive class labels & Class index or one-hot vector \\
\midrule
MultiLabelProcessor & Multiple simultaneous labels & Binary vector for each class \\
\midrule
RegressionLabelProcessor & Continuous target values & Scalar or vector of floats \\
\bottomrule
\end{tabularx}
\end{table*}

\subsection*{Specialized Processors}

Specialized processors address unique requirements of specific models or data formats. StageNet processors create tensor representations with stage-aware features for disease progression modeling. Multi-hot processors encode presence/absence patterns efficiently. Timeseries processors handle irregular temporal data with timestamps. Text processors prepare clinical notes and reports for language models, while raw processors pass data through unchanged for custom handling.

\begin{table*}[htbp]
\centering
\caption{Specialized Processors in PyHealth}
\label{tab:specialized_processors}
\begin{tabularx}{\textwidth}{l Y Y}
\toprule
\textbf{Processor} & \textbf{Description} & \textbf{Use Cases} \\
\midrule
StageNetProcessor & Stage-aware ICD code processing & StageNet model with disease progression \\
\midrule
StageNetTensorProcessor & Tensor representation for StageNet & Laboratory values with stage awareness \\
\midrule
MultiHotProcessor & Binary vector encoding of presence & Efficient code set representation \\
\midrule
TimeseriesProcessor & Irregular time series with timestamps & Vital signs, lab values with variable sampling \\
\midrule
TextProcessor & Clinical text preprocessing & Discharge summaries, radiology reports \\
\midrule
TensorProcessor & Generic tensor handling & Pre-processed numerical features \\
\midrule
RawProcessor & Pass-through without transformation & Custom preprocessing pipelines \\
\midrule
IgnoreProcessor & Placeholder for unused fields & Excluding fields from processing \\
\bottomrule
\end{tabularx}
\end{table*}

\section{PyHealth Interpretability Methods} \label{appdx: interpretability}

PyHealth implements a comprehensive suite of interpretability methods that explain model predictions through feature attribution and visualization techniques. These methods help clinicians and researchers understand which input features (e.g., specific diagnoses, lab values, or image regions) contribute most to predictions, enabling model validation, bias detection, and clinical insight discovery. The interpretability module supports both gradient-based and perturbation-based approaches across different data modalities, addressing the critical need for explainable AI in healthcare applications \cite{cina2025we_need_XAI}.

\begin{table}[htbp]
\centering
\caption{Overview of PyHealth Interpretability Methods}
\label{tab:interpretability_overview}
\begin{tabularx}{\textwidth}{lcp{0.6\textwidth}}
\toprule
\textbf{Method Category} & \textbf{Count} & \textbf{Description} \\
\midrule
Gradient-Based & 4 & Attribution through backpropagation of gradients \\
\midrule
Perturbation-Based & 2 & Attribution through input perturbations \\
\midrule
Attention-Based & 1 & Attribution from transformer attention mechanisms \\
\midrule
Visualization Tools & 4+ & Utilities for displaying and overlaying attributions \\
\bottomrule
\end{tabularx}
\end{table}

\subsection*{Gradient-Based Attribution Methods}

Gradient-based methods compute feature importance by analyzing how model outputs change with respect to input features \cite{simonyan2014saliency}. These methods are computationally efficient and work well with differentiable models. Basic gradient saliency maps provide first-order approximations, while integrated gradients \cite{sundararajan2017integrated} and DeepLift \cite{shrikumar2017deeplift} offer more sophisticated attributions that satisfy desirable theoretical properties like completeness and sensitivity.

\begin{table*}[htbp]
\centering
\caption{Gradient-Based Interpretability Methods in PyHealth}
\label{tab:gradient_methods}
\begin{tabularx}{\textwidth}{l Y Y}
\toprule
\textbf{Method} & \textbf{Description} & \textbf{Key Properties} \\
\midrule
BasicGradientSaliencyMaps  \cite{simonyan2014saliency} & Computes input gradient magnitude & Fast, first-order approximation \\
\midrule
IntegratedGradients \cite{sundararajan2017integrated} & Path integral of gradients from baseline & Satisfies completeness axiom \\
\midrule
DeepLift \cite{shrikumar2017deeplift}& Backpropagates contributions relative to reference & Handles saturation, efficient \\
\midrule
GIM (Gradient Input Multiplication) \cite{edin2025gimimprovedinterpretabilitylarge_GIM} & Element-wise product of gradients and inputs & Highlights salient input regions \\
\bottomrule
\end{tabularx}
\end{table*}

\subsection*{Perturbation-Based Attribution Methods}

Perturbation-based methods assess feature importance by observing how predictions change when inputs are masked or modified. LIME \cite{ribeiro2016lime} builds local linear approximations around individual predictions, while SHAP \cite{lundberg2017unifiedapproachinterpretingmodel_shap} computes Shapley values that provide game-theoretic optimal attributions. These methods are model-agnostic and can provide more faithful explanations at the cost of increased computational requirements. 

\begin{table*}[htbp]
\centering
\caption{Perturbation-Based Interpretability Methods in PyHealth}
\label{tab:perturbation_methods}
\begin{tabularx}{\textwidth}{l Y Y}
\toprule
\textbf{Method} & \textbf{Description} & \textbf{Key Properties} \\
\midrule
LimeExplainer & Local Interpretable Model-agnostic Explanations & Model-agnostic, local fidelity \\
\midrule
ShapExplainer & Shapley Additive exPlanations values & Theoretically optimal, consistent \\
\bottomrule
\end{tabularx}
\end{table*}

\subsection*{Attention-Based Attribution Methods}

Attention-based methods leverage built-in attention mechanisms in transformer models to derive feature importance. Chefer relevance propagation specifically addresses how to properly propagate relevance through multi-layer transformers, combining attention weights with gradient information to provide accurate attributions for vision transformers and other attention-based architectures.

\begin{table*}[htbp]
\centering
\caption{Attention-Based Interpretability Methods in PyHealth}
\label{tab:attention_methods}
\begin{tabularx}{\textwidth}{l Y Y}
\toprule
\textbf{Method} & \textbf{Description} & \textbf{Key Properties} \\
\midrule
CheferRelevance \cite{chefer2021genericattentionmodelexplainabilityinterpreting_attngrad} & Transformer-specific relevance propagation & Handles multi-head attention, layer propagation \\
\bottomrule
\end{tabularx}
\end{table*}

\section{PyHealth Uncertainty Quantification} \label{appdx: uncertainty}

PyHealth provides post-hoc uncertainty quantification methods that improve the reliability and trustworthiness of model predictions \cite{he2025survey}. The calibration module addresses two complementary aspects: probability calibration adjusts predicted probabilities to better reflect true confidence levels, while prediction set methods construct set-valued predictions with statistical coverage guarantees. These techniques are crucial for deploying healthcare AI systems in clinical settings where miscalibration can lead to harmful decisions.

\begin{table}[htbp]
\centering
\caption{Overview of PyHealth Uncertainty Quantification Methods}
\label{tab:uncertainty_overview}
\begin{tabularx}{\textwidth}{lcp{0.6\textwidth}}
\toprule
\textbf{Method Category} & \textbf{Count} & \textbf{Description} \\
\midrule
Calibration Methods & 4 & Post-hoc probability calibration techniques \\
\midrule
Prediction Sets & 4 & Conformal prediction and set-valued classifiers \\
\bottomrule
\end{tabularx}
\end{table}

\subsection*{Probability Calibration Methods}

Probability calibration methods adjust model outputs to ensure that predicted probabilities accurately reflect empirical frequencies \cite{guo2017calibration}. Temperature scaling learns a single scalar parameter to recalibrate logits, histogram binning uses non-parametric binning strategies \cite{zadrozny2001calibration}, Dirichlet calibration employs matrix transformations with regularization \cite{kull2019dirichlet}, and KCal leverages kernel density estimation on learned embeddings for multiclass calibration \cite{lin2023kcal}. These methods require a held-out calibration set and can significantly improve clinical decision-making based on predicted probabilities.

\begin{table*}[htbp]
\centering
\caption{Probability Calibration Methods in PyHealth}
\label{tab:calibration_methods}
\begin{tabularx}{\textwidth}{l Y l Y}
\toprule
\textbf{Method} & \textbf{Description} & \textbf{Modes} & \textbf{Reference} \\
\midrule
TemperatureScaling & Scalar temperature parameter for logit scaling & binary, multiclass, multilabel & \cite{guo2017calibration} \\
\midrule
HistogramBinning & Non-parametric binning of predictions & binary, multiclass, multilabel & \cite{zadrozny2001calibration} \\
\midrule
DirichletCalibration & Matrix transformation with regularization & multiclass & \cite{kull2019dirichlet} \\
\midrule
KCal & Kernel density estimation on embeddings & multiclass & \cite{lin2023kcal} \\
\bottomrule
\end{tabularx}
\end{table*}

\subsection*{Prediction Set Methods}

Prediction set methods provide set-valued predictions with statistical guarantees on coverage or error rates. Instead of outputting a single class, these methods return a set of plausible classes that contains the true label with high probability. LABEL \cite{sadinle2019label} and SCRIB \cite{lin2022scrib} offer complementary approaches with marginal and class-conditional coverage, FavMac \cite{lin2023favmac} handles multilabel scenarios with cost control, and CovariateLabel \cite{tibshirani2019conformal, laghuvarapu2023codrug} extends conformal prediction to handle distribution shift. These methods enable more conservative but reliable predictions in high-stakes clinical applications.

\begin{table*}[htbp]
\centering
\caption{Prediction Set Methods in PyHealth}
\label{tab:predictionset_methods}
\begin{tabularx}{\textwidth}{l Y l Y}
\toprule
\textbf{Method} & \textbf{Description} & \textbf{Modes} & \textbf{Reference} \\
\midrule
LABEL & Least Ambiguous set-valued classifier & multiclass & \cite{sadinle2019label} \\
\midrule
SCRIB & Class-specific risk bounds with optimized thresholds & multiclass & \cite{lin2022scrib} \\
\midrule
FavMac & Value-maximizing sets with cost control & multilabel & \cite{lin2023favmac} \\
\midrule
CovariateLabel & Conformal prediction under covariate shift & multiclass & \cite{tibshirani2019conformal, laghuvarapu2023codrug} \\
\bottomrule
\end{tabularx}
\end{table*}





\end{document}